%% file: main.tex
\documentclass[10pt,twocolumn,letterpaper]{article}

\usepackage[pagenumbers]{cvpr}
\usepackage{adjustbox}
\usepackage{tabularx,makecell,multirow}
\usepackage{colortbl}
\usepackage{capt-of}
\usepackage{amssymb}
\usepackage{placeins}
\usepackage{needspace}

\definecolor{linkblue}{rgb}{0.21,0.49,0.74}
\colorlet{cvprblue}{linkblue}
\definecolor{tablemeanblue}{RGB}{229,240,250}
\definecolor{tablegroupgray}{gray}{0.42}
\usepackage[pagebackref,breaklinks,colorlinks,allcolors=linkblue]{hyperref}

\makeatletter
\g@addto@macro\@maketitle{%
  \noindent\begin{minipage}{\textwidth}
\centering
\captionsetup{type=figure,skip=5pt}
\includegraphics[width=0.99\textwidth]{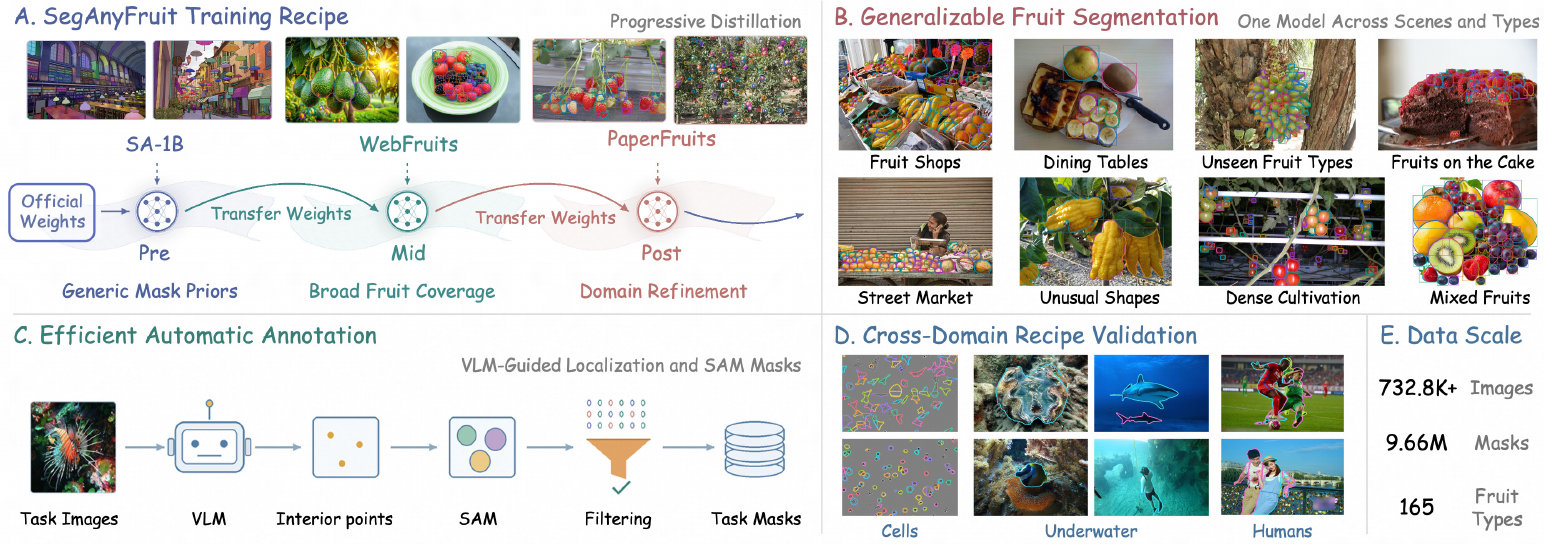}\par
\caption{\textbf{Overview of SegAnyFruit.}
(A) Three-stage \emph{Pre--Mid--Post} training recipe.
(B) Prompt-free fruit segmentation across diverse scenes and unseen types..
(C) Efficient automatic annotation.
(D) Cross-domain recipe validation.
(E) Scale of our curated resources.}
\label{fig:overview}
  \end{minipage}\par
  \vspace{18pt}
}
\makeatother

\begin{document}

\title{SegAnyFruit: Learning Generalizable and Edge Deployable\\
Fruit Instance Segmentation at Scale}

\author{%
  \adjustbox{max width=\textwidth}{%
    \normalsize
    \textbf{Yanan Wang}$^{1}$ \quad
    \textbf{Yuzheng Wang}$^{2}$ \quad
    \textbf{Yibin Ying}$^{1}$ \quad
    \textbf{Zhenghao Fei}$^{3}$\thanks{Corresponding author: Zhenghao Fei, Department of Biological and Agricultural Engineering, University of California, Davis, CA, United States.}%
  }\\[1mm]
  \adjustbox{max width=\textwidth}{%
    \small $^1$Zhejiang University \quad
    $^2$Alibaba Token Hub, Alibaba Group \quad
    $^3$University of California, Davis%
  }\\[1mm]
  {\small \texttt{\{mmwang@zju.edu.cn, zfei@ucdavis.edu\}}}
}
\maketitle

\begin{abstract}
Deploying fruit instance segmentation in real agricultural settings requires generalizing across unseen species, orchards, and imaging conditions under strict onboard compute and annotation budgets.
We introduce \textbf{SegAnyFruit} (SAF), a data construction and progressive training 
framework that distills foundation-model priors into lightweight, prompt-free segmenters. 
SAF combines generic mask pre-training, broad mid-training on automatically annotated web imagery from our curated \textbf{WebFruits} dataset, 
and human-verified agricultural refinement post-training on \textbf{PaperFruits}, together spanning 733K images and 9.66M whole-fruit masks across 165 fruit types. 
Across seven benchmarks, zero-shot \textbf{SAF-RF-DETR-XL} reaches 69.06 average mask AP$_{50:95}$ (+4.76 over SAM 3) 
and deployable on edge device. SAF initialization also boosts few-shot to full-shot adaptation across 
five agricultural domains while mitigating catastrophic forgetting. 
Finally, we systematically analyze training stages, data scale, model capacity, and generative augmentation, 
and validate our pipeline's transferability to underwater, cell and human segmentation. 
We will release our datasets, code, and SAF-models.
\end{abstract}

\noindent\textbf{Keywords:} Fruit instance segmentation; domain generalization; automatic annotation; synthetic data augmentation; edge deployment.

\section{Introduction}
\label{sec:intro}

Reliable fruit instance segmentation is critical for robotic harvesting, yield estimation, growth monitoring, and quality inspection. However, achieving robust instance segmentation across diverse visual environments with limited annotations remains a central challenge for agricultural perception. Fruits exhibit substantial variation in shape, color, size, and texture across varieties, species, growth stages, and cultivation practices, while dense foliage, occlusion, and changing illumination further complicate localization and instance separation~\cite{fei2021generalizable,jia2022polarnet}.

Closed-set specialist models, including YOLO-based architectures, have achieved strong in-domain accuracy on individual fruit benchmarks~\cite{smloyolo2026,zhang2026msrrt,tmaf2026,orange2026}. However, generalization to unseen orchards and unfamiliar fruit types remains challenging as acquisition conditions and visual distributions change~\cite{bargoti2017,metafruit2025}. Deploying these specialists in new settings therefore often requires renewed data collection, manual annotation, and fine-tuning.

Foundation models offer a promising alternative to dataset-specific training. Large vision-language models (VLMs), such as Qwen3-VL~\cite{bai2025qwen3vl}, and open-vocabulary detectors including OWLv2~\cite{minderer2023owlv2}, Grounding DINO~\cite{liu2024groundingdino}, DINO-X~\cite{ren2024dinox}, YOLO-World~\cite{cheng2024yoloworld}, and YOLOE~\cite{wang2025yoloe} extend recognition beyond fixed categories, while the Segment Anything family provides promptable instance segmentation~\cite{kirillov2023sam,carion2026sam3}. Recent fruit-focused efforts also build on foundation-model knowledge: FMFruit focuses on bounding-box detection~\cite{metafruit2025}, and SDM-D distills foundation-model supervision into compact students, using fruit-specific appearance descriptions during annotation~\cite{wang2026sdmd}. The computational demands, inference latency, and potential API dependence of frontier VLMs and large promptable segmenters further motivate efficient local inference on agricultural robots operating in the field.

In this work, we argue that achieving foundation-model-level generalization in a specialized perception task does not require a massive, general-purpose model. Instead, transferring foundation-model training principles, large-scale pretraining, diverse supervision, and task-specific refinement, can enable a compact segmenter to learn \emph{fruit} as a unified, generalizable visual concept across species and environments without target-specific prompts at inference.

To this end, we introduce \textbf{SegAnyFruit} (SAF), a data construction and progressive training framework for generalizable fruit instance segmentation (\Cref{fig:overview}). Starting from public pretrained weights, our SAF Training Recipe pretrains SAF models on large-scale general segmentation dataset SA-1B~\cite{kirillov2023sam} to acquire generic mask priors, mid-trains them on automatically annotated WebFruits to learn diverse fruit appearances and boundary cues, 
and post-trains them on human-calibrated PaperFruits from dense orchard environments to refine agricultural segmentation.
The SAF model family offers accuracy--speed trade-offs, including compact variants for edge deployment. Our SAF-RF-DETR-XL surpasses the evaluated VLMs and open-vocabulary baselines in average mask AP$_{50:95}$ across seven benchmarks. SAF initialization supports label-efficient adaptation and anti-forgetting across five unseen agricultural domains. 
Our contributions are fourfold:

\begin{itemize}
\interlinepenalty=1000
  \item \textbf{A progressive, cross-domain training recipe.}
  We propose a three-stage training recipe (Pre on large-scale general segmentation data, Mid on curated auto-labels, and Post on human-optimized annotations) that distills foundation-model priors into lightweight, prompt-free segmenters. Primarily validated on fruit perception, this training recipe also generalizes effectively to underwater, cell and human segmentation under scarce labels.
   \item \textbf{Large-scale, instance-curated fruit datasets.}
  We construct and will release WebFruits, PaperFruits, and seven manually annotated evaluation benchmarks, spanning 165 fruit types, 732,870 images, and 9.66M masks, paired with a category-aware curation pipeline enforcing whole-fruit instance granularity.
  \item \textbf{An open-source model family for zero-shot, adaptive, and edge deployment.}
  We release edge-friendly SAF-models across multiple architectures and scales. 
  Zero-shot SAF-RF-DETR-XL reaches 69.06 average mask AP$_{50:95}$ and SAF-YOLO26m-seg reaches 62.14, while their FP16 TensorRT engines run at 30~FPS and 69~FPS on Jetson Orin NX, respectively.
  Beyond zero-shot inference, SAF initialization improves few-shot to full-shot adaptation while preserving non-target performance.
  \item \textbf{Systematic study of supervision, scale, and label efficiency.}
  We analyze how training stages, data scale, model capacity, and generative video augmentation govern generalization and adaptation, offering practical guidelines for label-efficient domain segmentation.
\end{itemize}

\begin{figure*}[t]
\centering
\includegraphics[width=0.99\textwidth]{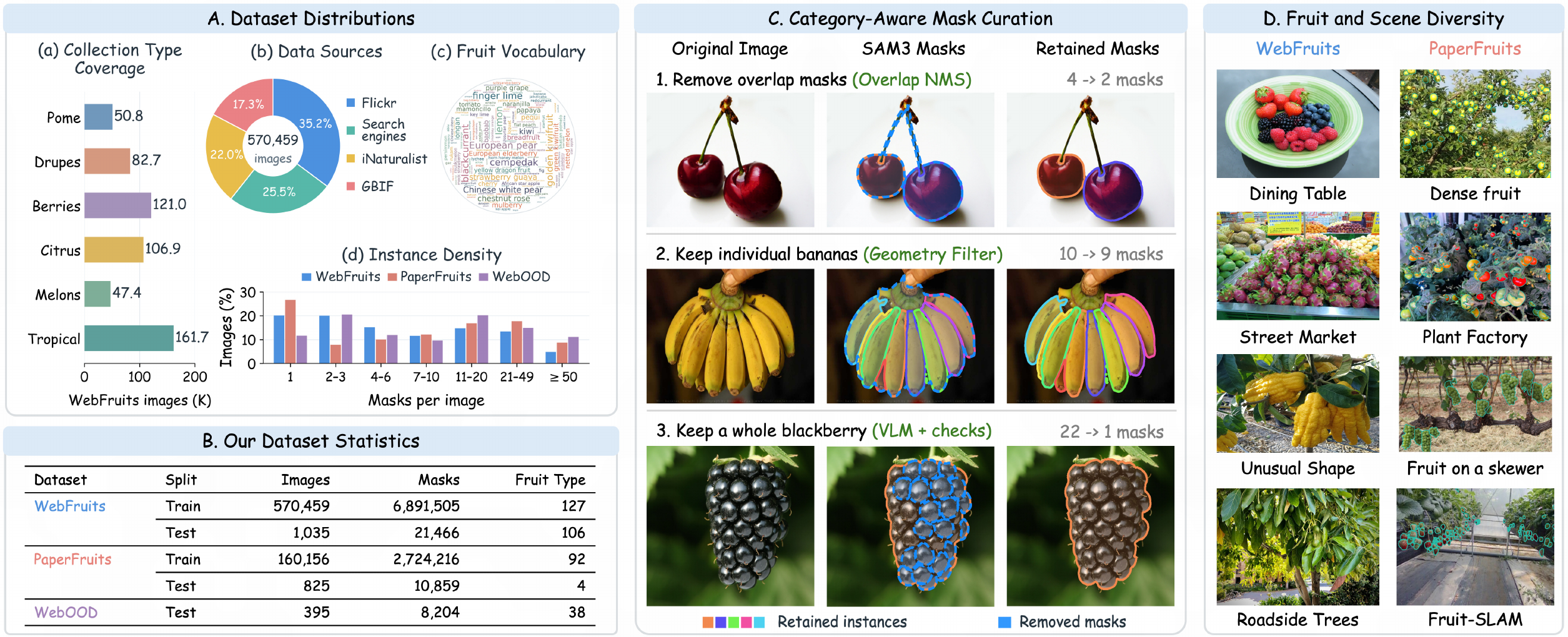}
\caption{WebFruits and PaperFruits. (A) Coverage and distributions.
(B) Image and mask counts; Train includes the internal validation split,
and PaperFruits Test summarizes the five held-out agricultural benchmarks.
Collection labels include varieties and visual forms.
(C) Whole-fruit mask curation. (D) Scene diversity.}
\label{fig:datasets}
\vspace{-6pt}
\end{figure*}

\vspace{-0.1em}
\section{Related Work}
\label{sec:related}
\vspace{-0.1em}
\textbf{Domain-specific visual models.}
Fruit perception has evolved from hand-crafted color and geometry heuristics to learned detectors and segmenters. DeepFruits and orchard Faster R-CNN~\cite{ren2015fasterrcnn} established deep multi-fruit detection~\cite{sa2016,bargoti2017}, Polar-Net improved mask boundaries under foliage occlusion~\cite{jia2022polarnet}, and recent YOLO variants prioritize real-time throughput~\cite{jocher2026yolo26}. Yet strong in-domain accuracy rarely transfers across orchards, imaging setups, or unseen species, and fixed classification heads prevent recognition beyond the training taxonomy~\cite{fei2021generalizable,liu2024groundingdino}. Rather than training per-dataset specialists, we learn \emph{fruit} as a unified visual concept that generalizes across scenes and unseen fruit types at field-ready latency.

\par\vspace{1.5pt}
\textbf{Foundation models and open-vocabulary perception.}
VLMs such as Qwen ground instructions into boxes and points~\cite{bai2025qwen3vl}, Grounding DINO enables language-guided localization~\cite{liu2024groundingdino}, and the Segment Anything family provides promptable instance segmentation~\cite{kirillov2023sam,ravi2024sam2,carion2026sam3}. While these frontier models generalize broadly, their heavy compute and latency prohibit real-time deployment on onboard agricultural hardware, and VLMs often lack mask boundary precision in dense foliage. Lighter open-vocabulary architectures such as YOLO-World, YOLOE, and EfficientViT-SAM improve inference efficiency~\cite{cheng2024yoloworld,wang2025yoloe,zhang2024efficientvitsam}, yet remain reliant on text prompts and underperform on small, heavily occluded fruits. In agriculture, FMFruit explores open-set multi-fruit detection but outputs bounding boxes rather than instance masks~\cite{metafruit2025}, while SDM-D distills SAM~2~\cite{ravi2024sam2} and OpenCLIP~\cite{cherti2023reproducible} into compact students but depends on fruit-specific appearance descriptions~\cite{wang2026sdmd}. Unlike these approaches, we target a compact, prompt-free fruit segmenter that delivers accurate instance masks across diverse environments while running locally on edge hardware.

\par\vspace{1.5pt}
\textbf{Representation learning and synthetic data.}
Strong visual representations reduce reliance on exhaustive manual labels. DINOv3 learns transferable dense features via large-scale self-supervision~\cite{simeoni2025dinov3} for downstream segmentation. Similarly, Depth Anything and DeFM show that combining broad pseudo-labeled supervision with distillation yields compact, generalizable models for robotic perception~\cite{yang2024depthanything,patel2026defm}, extending Noisy Student's self-training~\cite{xie2020noisystudent}. In parallel, generative augmentation offers complementary visual coverage: DiverGen scales synthetic diversity for instance segmentation~\cite{fan2024divergen}, and GAN-based crop simulation aids fruit-domain transfer~\cite{fei2021generalizable}. Building on these insights, we systematically examine how self-supervised and generic mask priors, curated web pseudo-labels, human-optimized agricultural data, and prompted video synthesis interact across model scales and label budgets.

\section{Data and Model Construction}
\label{sec:method}
To learn a prompt-free segmenter that generalizes across fruit species and environments, We curate two complementary corpora, WebFruits for broad visual diversity and PaperFruits for consolidated agricultural benchmarks with higher-fidelity supervision (\Cref{fig:datasets}), and train compact instance segmentation models through a progressive three-stage recipe (\textit{Pre}, \textit{Mid}, and \textit{Post})

\subsection{Data Collection}
\label{sec:fruit-data}
\paragraph{WebFruits and WebOOD.}
Using a unified web-crawling pipeline, we harvest $\sim$1.68M candidate images from Flickr~\cite{flickrapi}, iNaturalist~\cite{inaturalistapi}, GBIF~\cite{gbifoccurrenceapi}, and web search engines (\Cref{fig:datasets}A(b)). Search queries combine multilingual common/scientific names and varieties with \emph{cultivation contexts} (e.g., orchards, greenhouses, hydroponics, harvesting) and \emph{appearance states} (e.g., unripe, overripe, diseased, cross-sectional views), such as ``Malus domestica,'' ``apple picking farm,'' and ``unripe green apple.'' To ensure strict vocabulary separation, we partition query terms into two disjoint sets and verify separation via image metadata analysis alongside readability checks, perceptual-hash deduplication, and mask curation. This yields \textbf{WebFruits} with 571,494 images (570,459 for training/validation and 1,035 held out as \emph{WebTest}), and \textbf{WebOOD} with 395 manually annotated images spanning 38 unseen fruit types strictly excluded from training and prompt selection.

\textbf{PaperFruits.}
We consolidate 36 agricultural and classification datasets, reserving five agricultural benchmarks for evaluation and 31 sources for training and validation. Representative sources include WGISD~\cite{santos2020wgisd}, MinneApple~\cite{hani2020minneapple}, ACFR~\cite{bargoti2017}, and VegFru~\cite{hou2017vegfru}, combining field imagery with broad fruit appearance diversity. PaperFruits comprises 160,981 images spanning orchards, greenhouses, robot-captured scenes, and classification collections (160,156 train/val and 825 test; \Cref{fig:datasets}D).

\subsection{Data Annotation and Quality Control}
\label{sec:data-annotation}
\paragraph{WebFruits: scalable automatic annotation.}
We annotate WebFruits using SAM~3~\cite{carion2026sam3}. For each fruit collection, we evaluate three prompt formats on 50 randomly sampled images: the generic label (``fruit''), the specific category name (e.g., ``apple''), and their union (``apple or fruit''). We then select the format that achieves the best coverage and instance granularity for batch annotation.

\textbf{Category-aware whole-fruit mask curation.}
Raw SAM~3 proposals often duplicate instances, merge fruit clusters, or over-segment surface parts. We apply a three-step curation pipeline without altering retained mask boundaries (\Cref{fig:datasets}C):
(1)~\emph{Duplicate suppression (Overlap NMS)} retains only the higher-confidence mask when two proposals share $>90\%$ intersection over each mask's area;
(2)~\emph{Group suppression (Geometry Filter)} discards a parent mask if it contains at least two retained children (each $>90\%$ inside the parent) whose union covers $>60\%$ of the parent's area—relaxed to one child for banana/plantain and tomato/cherry-tomato to prevent cluster merging; and
(3)~\emph{Semantic selection (VLM + checks)} resolves compound fruits such as blackberries by querying a VLM to retain whole-fruit masks (confidence $\geq 0.75$) while rejecting individual drupelets, clusters, and background regions.
Grape, raspberry, and mulberry undergo only duplicate suppression to avoid discarding valid cluster units. Images yielding no retained masks are excluded.

\textbf{PaperFruits: human-optimized annotation.}
The 31 training sources from PaperFruits follow four annotation routes tailored to their native labels: 
3 datasets (WGISD, Berremangra, Dyson RGB)~\cite{santos2020wgisd,beldek2026berremangra,tafuro2022strawberry} retain native masks; 9 box-labeled datasets (e.g., Litchi-UAV, Oranges Field, AgRobTomato)~\cite{li2026litchiuav,lucas2025orangesfield,magalhaes2021agrobtomato} prompt SAM~3 with original boxes; 
18 classification/box-free sources (e.g., FruitBench, FruitVision, VegFru)~\cite{tjietfruitbench,bijoy2025fruitvision,hou2017vegfru} use text-prompted SAM~3; and Schram03~\cite{schram03fruits} uses boxes and points derived from foreground priors. 
Six mask/box sources are further supplemented with text-prompted proposals.
To broaden coverage, we merge proposals from multiple prompts, filter them by confidence, overlap, and area limits while protecting native masks. After a final pass to remove near-duplicates (details in the supplement), we manually inspect a sampled subset of labeled data to prevent systematic errors.

All test images are manually annotated.

\subsection{Dataset Statistics and Splits}
\label{sec:data-statistics}
\paragraph{Training corpora.}
As summarized in \Cref{fig:datasets}B, the WebFruits train/val pool comprises 570,459 images and 6,891,505 masks across 127 collection types, while PaperFruits comprises 160,156 images and 2,724,216 masks across 92 types (spanning 165 total fruit types including varieties and visual forms, \Cref{fig:datasets}A(c)). WebFruits spans six major botanical/culinary groups, namely pome, drupes, berries, citrus, melons/fruit vegetables, and tropical/other fruits (\Cref{fig:datasets}A(a)), across dining tables, markets, roadside trees, and diverse ripeness states (\Cref{fig:datasets}A(b), D). PaperFruits emphasizes dense agricultural cultivation. Both corpora span sparse and crowded scenes (\Cref{fig:datasets}A(d)), averaging 12.1 and 17.0 instances per image (with 4.9\% and 8.8\% of images containing $\geq 50$ masks), providing strong supervision for separating touching instances.

\textbf{Evaluation benchmarks.}
We evaluate across seven manually annotated test sets spanning three generalization regimes (\Cref{fig:datasets}B). \emph{WebTest} measures held-out performance within the WebFruits vocabulary, whereas \emph{WebOOD} tests zero-shot transfer to 38 unseen fruit species. Between these regimes, five agricultural datasets, strictly excluded from training and model selection, probe real-world domain shifts: \emph{MSU-Apple}~\cite{chu2023o2rnet} captures clustered apples under daylight foliage occlusion; \emph{KFuji RGB-DS}~\cite{genemola2019kfuji} images apples under artificial nighttime lighting to test intra-class acquisition shifts; \emph{MangoFruitBD}~\cite{islam2026mangofruitbd} spans healthy and diseased mangoes; \emph{StrawDI\_Db1}~\cite{perez2020strawdi} features small, unripe, and occluded plantation strawberries; and \emph{Rob2Pheno Tomato}~\cite{afonso2020tomato} averages 78.3 tomatoes per nighttime greenhouse image to stress ultra-dense instance separation.

\subsection{Model-Agnostic Progressive Training Recipe}
\label{sec:training-recipe}
Beyond the fruit datasets, our SAF Training Recipe embodies a generalizable, architecture-agnostic paradigm for label-efficient domain segmentation (\Cref{fig:overview}A). Its core principle is to decouple supervision into three progressive stages:
(1)~\emph{Pre-training (Pre)} on generic segmentation data builds class-agnostic objectness and boundary priors;
(2)~\emph{Mid-training (Mid)} leverages foundation models (VLMs and SAM) to automatically annotate diverse in-the-wild imagery, replacing expensive manual labeling with scalable, curated pseudo-supervision; and
(3)~\emph{Post-training (Post)} refines dense instance separation and domain-specific boundaries using a compact set of human-calibrated annotations.

In our fruit instantiation, these three stages correspond to SA-1B~\cite{kirillov2023sam} (\emph{Pre}), WebFruits (\emph{Mid}), and PaperFruits (\emph{Post}). Crucially, the recipe imposes no custom network modules or specialized losses—training standard detectors directly through their native objectives. We instantiate it across three distinct architectural families: CNN-based YOLO26-seg~\cite{jocher2026yolo26}, DETR-style RF-DETR~\cite{robinson2026rfdetr}, and frozen DINOv3-B/16~\cite{simeoni2025dinov3} with Mask2Former~\cite{cheng2022mask2former}. Foundation-model teachers and text prompts are used only during data construction; compact SAF models perform fast, prompt-free inference.

\section{Experiments}
\label{sec:experiments}
\label{sec:results}

\begin{table*}[t]
\centering
\captionsetup{skip=3pt}
\caption{Fruit segmentation without target-domain adaptation.
Entries: mask AP$_{50:95}$/AP$_{50}$. Avg. AP$_{50:95}$ is the equal-weight average of mask AP$_{50:95}$ across the seven datasets.
The six tests other than WebTest are OOD relative to our fruit training.
$\ell$ denotes the long side. $\dagger$VLM inputs use native resolution, downscaled to $\ell=1280$ only when $\ell>1280$.
Best and second best are \textbf{bold} and \underline{underlined}.}
\label{tab:main}
\begingroup
\fontsize{7.5}{9.5}\selectfont
\setlength{\tabcolsep}{1.2pt}
\renewcommand{\arraystretch}{1.0}
\renewcommand{\tabularxcolumn}[1]{m{#1}}
\newcommand{\tblmodel}[1]{\makecell[l]{#1}}
\begin{tabularx}{\textwidth}{@{}m{78pt}>{\centering\arraybackslash}m{33pt}*{7}{>{\centering\arraybackslash}X}>{\columncolor{tablemeanblue}\centering\arraybackslash}m{48pt}@{}}
\toprule
Model & Input & WebTest & MSU-Apple & KFuji-Apple & Mango & StrawDI & Tomato & WebOOD & Avg. AP$_{50:95}$ \\
\midrule
\multicolumn{10}{@{}l}{\color{tablegroupgray}\itshape General-Purpose Vision--Language Models} \\[1pt]
\tblmodel{GPT-6-Astra~\cite{openai2026astra}} & Native$^{\dagger}$ & 41.30\,/\,69.19 & 33.22\,/\,63.32 & 25.83\,/\,53.06 & 84.74\,/\,96.01 & 58.58\,/\,81.39 & 37.84\,/\,72.45 & 49.60\,/\,75.69 & 47.30 \\
\tblmodel{GPT-6.1-Terra~\cite{openai2026terra}} & Native$^{\dagger}$ & 31.17\,/\,52.73 & 34.57\,/\,62.26 & 31.08\,/\,56.84 & 75.05\,/\,89.90 & 55.96\,/\,79.68 & 30.69\,/\,54.30 & 35.62\,/\,56.46 & 42.02 \\
\tblmodel{GPT-6.1-Sol~\cite{openai2026sol}} & Native$^{\dagger}$ & 38.72\,/\,66.93 & 29.32\,/\,59.09 & 25.14\,/\,52.93 & 81.90\,/\,94.84 & 57.01\,/\,80.16 & 33.08\,/\,66.12 & 45.80\,/\,73.18 & 44.42 \\
\tblmodel{GPT-6-Luna~\cite{openai2026luna}} & Native$^{\dagger}$ & 7.14\,/\,16.91 & 8.29\,/\,22.69 & 9.21\,/\,24.48 & 57.93\,/\,88.45 & 37.51\,/\,67.41 & 4.90\,/\,13.43 & 8.63\,/\,19.01 & 19.09 \\
\tblmodel{DeepSeek-V4.1~\cite{deepseek2026v41}} & Native$^{\dagger}$ & 3.62\,/\,9.07 & 3.39\,/\,11.69 & 2.52\,/\,9.86 & 50.25\,/\,87.54 & 24.98\,/\,57.81 & 2.58\,/\,8.31 & 5.73\,/\,14.28 & 13.30 \\
\tblmodel{GLM-5.3-Flash~\cite{zai2026glm53flash}} & Native$^{\dagger}$ & 12.19\,/\,32.41 & 8.89\,/\,27.03 & 4.95\,/\,18.28 & 59.71\,/\,91.13 & 27.51\,/\,61.05 & 7.13\,/\,16.33 & 8.07\,/\,19.79 & 18.35 \\
\midrule
\multicolumn{10}{@{}l}{\color{tablegroupgray}\itshape Open-Vocabulary Segmentation Models} \\[1pt]
\tblmodel{YOLOE~\cite{wang2025yoloe}} & $640^2$ & 20.98\,/\,27.52 & 26.61\,/\,38.72 & 16.00\,/\,21.41 & 83.53\,/\,90.60 & 26.97\,/\,40.48 & 24.06\,/\,33.58 & 31.48\,/\,39.48 & 32.80 \\
\tblmodel{YOLO-World~\cite{cheng2024yoloworld}} & $640^2$ & 23.97\,/\,40.03 & 36.94\,/\,60.44 & 29.62\,/\,42.69 & 61.46\,/\,80.40 & 27.67\,/\,45.20 & 24.21\,/\,37.42 & 29.93\,/\,49.26 & 33.40 \\
\tblmodel{Grounded SAM 2~\cite{groundedsam2}} & $1024^2$ & 44.06\,/\,56.85 & 47.08\,/\,67.96 & 43.93\,/\,67.63 & 92.13\,/\,\textbf{98.60} & 59.02\,/\,77.94 & 63.35\,/\,78.67 & 50.37\,/\,60.99 & 57.13 \\
\tblmodel{SAM 3~\cite{carion2026sam3}} & $644^2$ & 51.59\,/\,68.19 & 55.27\,/\,\underline{80.53} & 36.98\,/\,60.11 & 89.06\,/\,95.75 & 60.11\,/\,81.55 & 60.92\,/\,82.67 & 61.31\,/\,75.75 & 59.32 \\
\tblmodel{SAM 3~\cite{carion2026sam3}} & $1008^2$ & \underline{59.50}\,/\,74.65 & \underline{60.10}\,/\,\textbf{81.09} & 35.36\,/\,58.06 & 90.49\,/\,96.73 & 66.31\,/\,83.57 & \textbf{74.07}\,/\,\textbf{89.02} & \underline{64.30}\,/\,\underline{79.56} & 64.30 \\
\midrule
\multicolumn{10}{@{}l}{\bfseries SAF (Ours)} \\[1pt]
\tblmodel{SAF-YOLO26m-seg} & $640^2$ & 53.94\,/\,\underline{75.24} & 49.33\,/\,76.43 & \textbf{49.45}\,/\,\underline{74.45} & 92.25\,/\,\underline{97.28} & 70.29\,/\,\underline{88.75} & 57.48\,/\,80.86 & 62.28\,/\,79.39 & 62.14 \\
\tblmodel{SAF-DINOv3+M2F} & $640^2$ & 55.19\,/\,71.39 & 52.20\,/\,74.97 & 46.20\,/\,71.88 & 92.34\,/\,96.26 & 68.25\,/\,81.91 & 56.32\,/\,73.93 & 60.87\,/\,77.40 & 61.62 \\
\tblmodel{SAF-DINOv3+M2F} & $\ell{\leq}1024$ & 58.86\,/\,74.57 & 56.57\,/\,76.80 & 46.64\,/\,71.72 & \underline{92.69}\,/\,96.43 & \underline{71.34}\,/\,84.25 & 65.79\,/\,80.93 & 63.32\,/\,79.24 & \underline{65.03} \\
\tblmodel{SAF-RF-DETR-XL} & $624^2$ & \textbf{64.01}\,/\,\textbf{79.51} & \textbf{60.14}\,/\,80.46 & \underline{48.39}\,/\,\textbf{75.49} & \textbf{94.21}\,/\,97.06 & \textbf{75.33}\,/\,\textbf{90.03} & \underline{72.95}\,/\,\underline{87.47} & \textbf{68.38}\,/\,\textbf{82.66} & \textbf{69.06} \\
\bottomrule
\end{tabularx}
\endgroup
\end{table*}

\paragraph{Evaluation protocol and baselines.}
We evaluate models on seven manually annotated benchmarks from \Cref{sec:data-statistics}, defining OOD relative to our fruit training collections and vocabulary. Following COCO evaluation~\cite{lin2014coco}, we report mask AP$_{50:95}$ and mask AP$_{50}$. We compare SAF against general-purpose VLMs and open-vocabulary segmenters using their strongest of three tested prompts, whereas SAF runs prompt-free. We report native input resolutions and provide box metrics in the supplement.

\subsection{Generalization and Edge Deployment}
\label{sec:main-results}

\begin{figure}[!t]
\centering
\includegraphics[trim=0 45.6bp 0 0,clip,width=\columnwidth]{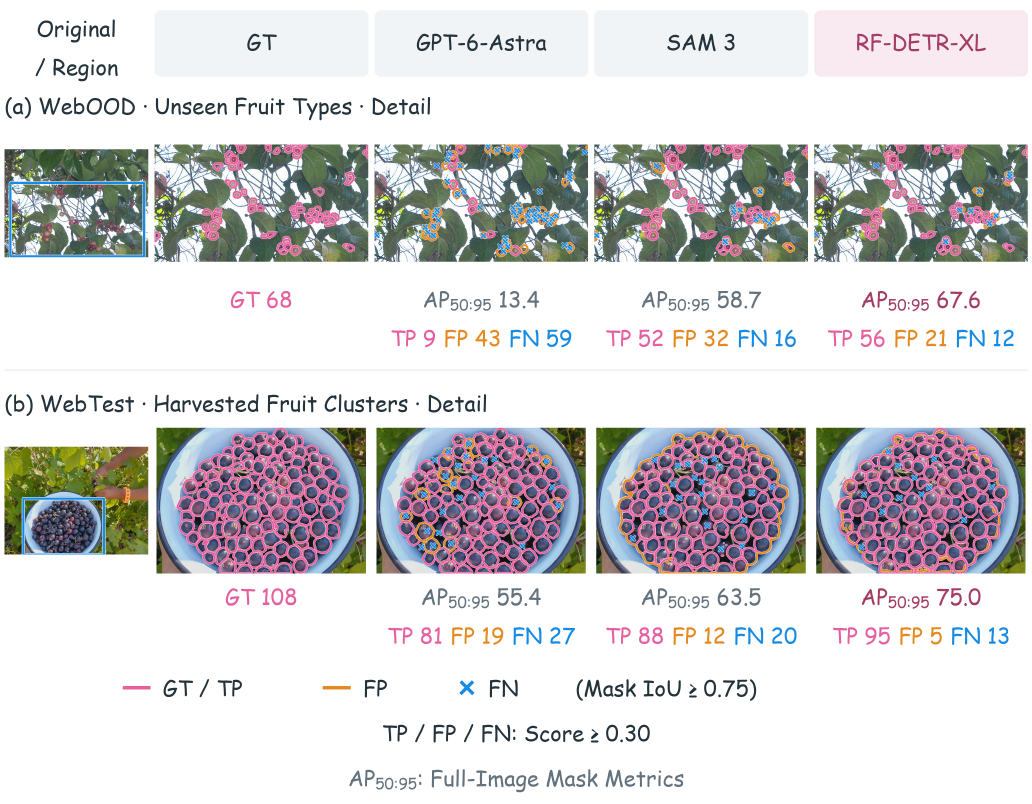}
\captionsetup{skip=3pt}
\caption{Qualitative comparison with SAF-RF-DETR-XL on (a) unseen fruit types in WebOOD and (b) clustered blackcurrants.}
\label{fig:blackcurrant-comparison}
\vspace{-14pt}
\end{figure}

\textbf{Zero-shot accuracy across species and field conditions.}
All four SAF configurations outperform Grounded SAM~2 and SAM~3 ($644^2$) in \Cref{tab:main}, reaching up to 69.06 average mask AP$_{50:95}$ (+4.76 over SAM~3 at $1008^2$), with SAF-RF-DETR-XL and SAF-DINOv3+M2F ($\ell{\leq}1024$) setting the top marks. Across the seven benchmarks, SAF models rank first on six (five led by SAF-RF-DETR-XL, including a +9.02 AP$_{50:95}$ gain on StrawDI, and KFuji-Apple led by SAF-YOLO26m-seg) while remaining competitive with the 848M-parameter SAM~3 on Tomato. Beyond cross-species generalization, SAF resists intra-class acquisition shifts: moving from daylight MSU-Apple to nighttime KFuji-Apple degrades SAM~3 ($1008^2$) by 24.74 AP$_{50:95}$, whereas SAF-RF-DETR-XL cuts this drop by more than half (to 11.75) and SAF-YOLO26m-seg maintains virtually unchanged accuracy. Qualitatively (\Cref{fig:blackcurrant-comparison}), SAF-RF-DETR-XL produces cleaner boundaries with fewer false positives on unseen WebOOD fruits and harvested clusters. Together, these results show that distilling intrinsic morphological priors yields far stronger perceptual invariance than relying on generic text-prompted foundation models.

\textbf{Mask precision separates SAF from VLMs.}
Evaluated frontier VLMs trail even our lightest SAF model by at least 14.84 average mask AP$_{50:95}$ (\Cref{tab:main}). While GPT-6-Astra reaches 84.74 AP$_{50:95}$ on relatively isolated Mango fruits, its mask accuracy drops sharply to 37.84 on dense, occluded Tomato canopies. Visually, VLMs suffer from severe false negatives and coarse contours in crowded clusters (\Cref{fig:blackcurrant-comparison}a), confirming that broad semantic grounding cannot replace dedicated instance boundary supervision.

\begin{table}[t]
\centering
\captionsetup{skip=3pt}
\caption{TensorRT engine throughput on Jetson Orin NX 8GB.}
\label{tab:edge-deployment}
\begingroup
\fontsize{8}{10}\selectfont
\setlength{\tabcolsep}{2pt}
\renewcommand{\arraystretch}{1.0}
\begin{tabularx}{\columnwidth}{@{}
>{\hsize=1.5\hsize\linewidth=\hsize\raggedright\arraybackslash}X
*{2}{>{\hsize=0.95\hsize\linewidth=\hsize\centering\arraybackslash}X}
*{2}{>{\hsize=0.8\hsize\linewidth=\hsize\centering\arraybackslash}X}@{}}
\toprule
Model & \mbox{Params (M)} & Input & \mbox{FP32 FPS} & \mbox{FP16 FPS} \\
\midrule
SAM3 & 848 & $1008^2$ & 1.40 & 1.86 \\
\midrule
YOLO26n-seg & 3.1 & $640^2$ & 121.75 & 253.75 \\
YOLO26s-seg & 11.4 & $640^2$ & 64.90 & 148.83 \\
YOLO26m-seg & 27.0 & $640^2$ & 28.79 & 69.22 \\
\midrule
DINOv3+M2F & 116.1 & $480\!\times\!640$ & 3.33 & 5.51 \\
\midrule
RF-DETR-M & 35.4 & $432^2$ & 23.04 & 66.87 \\
RF-DETR-L & 35.9 & $504^2$ & 16.21 & 49.51 \\
RF-DETR-XL & 38.1 & $624^2$ & 8.52 & 30.01 \\
\bottomrule
\end{tabularx}
\endgroup
\vspace{-2pt}
\end{table}

\begin{figure*}[!t]
\centering
\captionsetup{skip=-4pt}
\includegraphics[width=\textwidth]{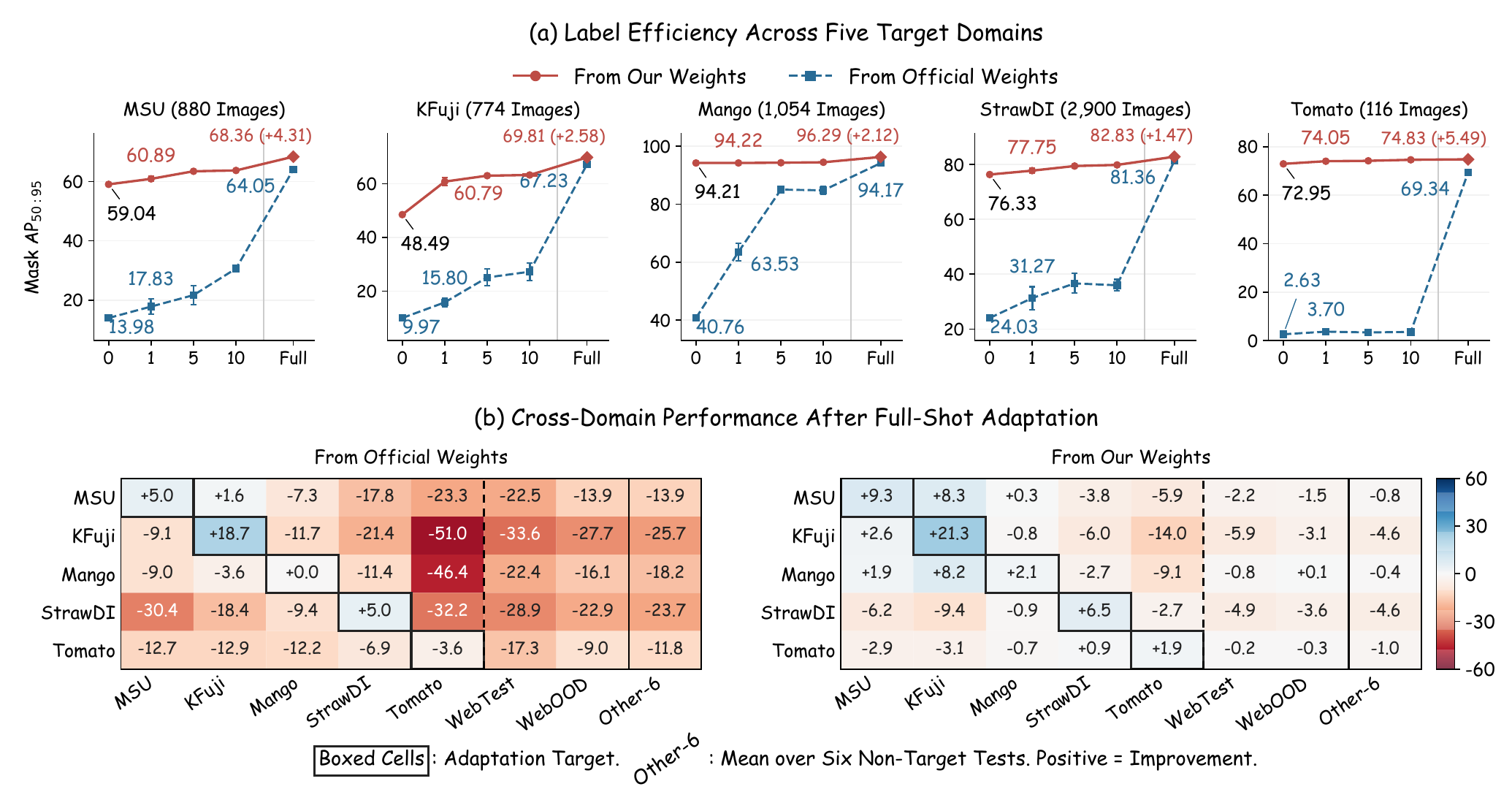}
\caption{Adaptation and retention with RF-DETR-XL.
(a) Target AP$_{50:95}$ vs. labeled images (error bars: sample SD over five runs).
(b) Full-shot AP$_{50:95}$ changes over the unadapted SAF checkpoint.
Boxes mark adaptation targets; Other-6 averages non-target tests.}
\label{fig:fewshot-retention}
\vspace{-12pt}
\end{figure*}

\textbf{Pixel efficiency and real-time edge deployment.}
Higher input resolution consistently improves dense fruit segmentation, lifting DINOv3+M2F and SAM~3 by 3.41 and 4.98 average mask AP$_{50:95}$, respectively. However, SAF models achieve strong accuracy without relying on high-resolution inputs. Operating at $640^2$ ($\sim$60\% fewer pixels than SAM~3 at $1008^2$), SAF-YOLO26m-seg still outperforms high-resolution SAM~3 on KFuji, Mango, and StrawDI.
On a Jetson Orin NX 8GB (Super Mode, \Cref{tab:edge-deployment}), heavy foundation backbones fall far short of real-time speed, with SAM~3 and DINOv3+M2F reaching only 1.86 and 5.51 TensorRT~\cite{nvidia2026tensorrt} FP16 FPS. In contrast, all SAF detectors run comfortably in real time: RF-DETR variants deliver 30.01 to 66.87 FPS (making SAF-RF-DETR-XL over $16\times$ faster than SAM~3 with $22\times$ fewer parameters), while YOLO26-seg scales up to 253.75 FPS. Additional edge benchmarks are detailed in the supplement.

\subsection{Few-shot adaptation and retention}
\label{sec:adaptation}

We evaluate SAF's target adaptation and broad retention across 1-, 5-, 10-, and full-shot settings (1-shot: one labeled image) over five seeds, comparing RF-DETR-XL initialized from SAF versus official pretrained weights (\Cref{fig:fewshot-retention}; details in the supplement).

\textbf{Label-efficient adaptation and higher performance ceiling.}
SAF provides a strong domain-aligned initialization across all five agricultural benchmarks (\Cref{fig:fewshot-retention}a). Without any target supervision, zero-shot SAF already matches the full-shot official-initialized specialist on Mango and surpasses it by 3.61 AP$_{50:95}$ on dense Tomato canopies. Under extreme label scarcity, fine-tuning on a single image triggers rapid adaptation, enabling SAF to outperform 1-shot official baselines by +44.99 AP$_{50:95}$ on KFuji and +43.06 on MSU. Crucially, these advantages persist as target data scales: under full supervision, SAF initialization consistently raises the performance ceiling across all five domains (+1.47 to +5.49 AP$_{50:95}$ over official weights), showing that agricultural pre-training improves both sample efficiency and asymptotic accuracy.

\textbf{Preserving cross-environment generalization after specialization.}
\Cref{fig:fewshot-retention}b compares post-adaptation AP$_{50:95}$ across all seven benchmarks relative to the same unadapted SAF checkpoint. Standard full-shot training from official weights leaves severe cross-environment deficits, trailing unadapted SAF by 11.8--25.7 AP$_{50:95}$ on average across the six non-target benchmarks (\textit{Other-6}). In contrast, adapting from SAF improves every target domain (+1.9 to +21.3 AP$_{50:95}$) while keeping the average \textit{Other-6} reduction within just 0.4--4.6 AP$_{50:95}$. Notably, SAF even enables positive cross-environment transfer: adapting to MSU or Mango raises KFuji performance by +8.3 and +8.2 AP$_{50:95}$, respectively. Even under the sharpest cross-species shift (KFuji to Tomato), SAF limits the off-target drop to 14.0 AP$_{50:95}$ (versus 51.0 points below unadapted SAF for the official-initialized specialist), demonstrating strong resistance to catastrophic forgetting.

\subsection{Supervision budgets and model scale}
\label{sec:data-analysis}
\begin{figure}[t]
\centering
\includegraphics[width=\columnwidth]{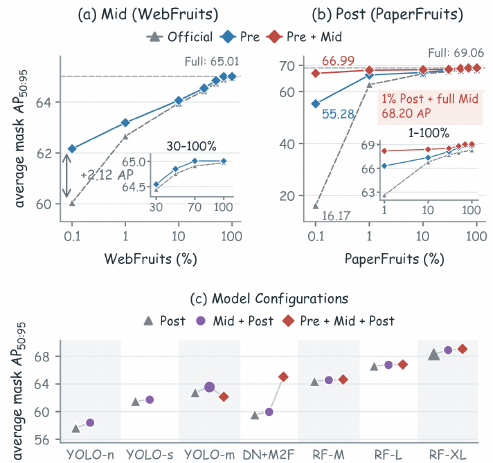}
\captionsetup{skip=5pt}
\caption{Data and model scaling. Seven-test average mask AP$_{50:95}$ for
(a) WebFruits Mid, (b) PaperFruits Post with three initializations,
and (c) model configurations.}
\label{fig:budget}
\vspace{-5pt}
\end{figure}

\Cref{tab:sources} compares full-data training routes across architectures, while \Cref{fig:budget} examines how performance scales with supervision budgets and model capacity.

\setcounter{topnumber}{1} %
\begin{table}[t]
\centering
\captionsetup{skip=3pt}
\caption{Training-stage ablation. Seven-test average mask AP$_{50:95}$;
$\checkmark$/$\times$: stage included/omitted. Column bests are bold.}
\label{tab:sources}
\begingroup
\fontsize{8}{10}\selectfont
\setlength{\tabcolsep}{0.5pt}
\renewcommand{\arraystretch}{1.0}
\begin{tabularx}{\columnwidth}{@{}
*{3}{>{\hsize=0.55\hsize\linewidth=\hsize\centering\arraybackslash}X}
>{\hsize=1.4\hsize\linewidth=\hsize\centering\arraybackslash}X
>{\hsize=1.8\hsize\linewidth=\hsize\centering\arraybackslash}X
>{\hsize=1.15\hsize\linewidth=\hsize\centering\arraybackslash}X@{}}
\toprule
Pre & Mid & Post & YOLO26m & \mbox{DINOv3+M2F} & RF-XL \\
\midrule
$\times$ & $\checkmark$ & $\times$ & 50.45 & 56.69 & 64.97 \\
$\checkmark$ & $\checkmark$ & $\times$ & 50.93 & 62.72 & 65.01 \\
\midrule
$\times$ & $\times$ & $\checkmark$ & 62.77 & 59.57 & 68.29 \\
$\checkmark$ & $\times$ & $\checkmark$ & 62.32 & 63.33 & 68.83 \\
$\times$ & $\checkmark$ & $\checkmark$ & \textbf{63.54} & 59.96 & 68.89 \\
$\checkmark$ & $\checkmark$ & $\checkmark$ & 62.14 & \textbf{65.03} & \textbf{69.06} \\
\bottomrule
\end{tabularx}
\endgroup
\vspace{-1pt}
\end{table}

\textbf{Dense cultivation supervision anchors final accuracy.}
Across all architectures, incorporating PaperFruits (\textit{Post}) yields the largest single-stage improvement, boosting results by 11.21, 2.31, and 4.05 average mask AP$_{50:95}$ on YOLO26m, DINOv3+M2F, and RF-DETR-XL, respectively (\Cref{tab:sources}). While WebFruits (\textit{Mid}) provides broad appearance coverage, PaperFruits supplies dense orchard scenes and refined instance boundaries essential for practical fruit segmentation. Upstream stages (\textit{Pre} and \textit{Mid}) contribute complementary priors whose impact varies by architecture: DINOv3+M2F gains 5.46 average mask AP$_{50:95}$ over Post-only training with all three stages (reaching 65.03 average mask AP$_{50:95}$), and RF-DETR-XL peaks at 69.06 average mask AP$_{50:95}$ with the full pipeline, whereas YOLO26m achieves its strongest result (63.54 average mask AP$_{50:95}$) with Mid~+~Post.

\textbf{Upstream supervision dramatically reduces downstream annotation needs.}
While upstream stages yield modest gains when Post data are abundant, they act as powerful annotation multipliers in label-scarce regimes (\Cref{fig:budget}a--b). During WebFruits scaling (\Cref{fig:budget}a), SA-1B Pre provides a +2.12 average mask AP$_{50:95}$ head start at 0.1\% data, after which both Pre and official-initialized curves steadily climb and converge at 100\% WebFruits. When transitioning to PaperFruits with a 0.1\% label pool (\Cref{fig:budget}b), adding full Mid on top of Pre raises performance from 55.28 to 66.99 average mask AP$_{50:95}$ (+11.71 across seven benchmarks), far surpassing the 16.17 from official weights. Pairing full Mid with just 1\% PaperFruits reaches 68.20 average mask AP$_{50:95}$, recovering 98.8\% of the full-data reference (69.06) with two orders of magnitude fewer manual labels, as Mid absorbs most domain variance and leaves Post to calibrate fine boundaries.

\textbf{Model capacity governs the ceiling of multi-stage supervision.}
\Cref{fig:budget}c shows accuracy improving with model scale within each family, compounded by higher native input resolutions in larger RF-DETR variants. With abundant Post supervision, compact models (SAF-YOLO26n/s and SAF-RF-DETR-M) saturate quickly and gain modestly from upstream training, whereas higher-capacity models (SAF-DINOv3+M2F and SAF-RF-DETR-XL) possess the representational bandwidth to retain web-scale priors and benefit from the full three-stage recipe. Consequently, compact students suit edge deployment when curated Post data are available, while larger models unlock the full value of large-scale upstream supervision.

\begin{table}[t]
\centering
\captionsetup{skip=3pt}
\caption{Synthetic augmentation with 320 human-labeled images.
+Syn adds 31,308 automatically labeled frames to the same originals.
All entries are mask AP$_{50:95}$. Average: equal-weight seven-test average; $\Delta$: +Syn $-$ Human.}
\label{tab:synthetic}
\begingroup
\fontsize{8}{10}\selectfont
\setlength{\tabcolsep}{2pt}
\renewcommand{\arraystretch}{1.0}
\begin{tabularx}{\columnwidth}{@{}l*{5}{>{\centering\arraybackslash}X}@{}}
\toprule
\multirow{2}{*}{Initialization} & \multicolumn{2}{c}{average mask AP$_{50:95}$} & \multicolumn{3}{c}{$\Delta$ AP$_{50:95}$} \\
\cmidrule(lr){2-3}\cmidrule(l){4-6}
 & Human & +Syn & Average & KFuji & WebOOD \\
\midrule
Official & 58.90 & \textbf{63.75} & $+4.85$ & $+4.26$ & $+8.41$ \\
Pre & 62.62 & \textbf{63.65} & $+1.04$ & $+4.62$ & $+0.32$ \\
Pre + Mid & \textbf{66.91} & 64.69 & $-2.22$ & $+2.47$ & $+0.68$ \\
\bottomrule
\end{tabularx}
\endgroup
\end{table}
\begin{figure}[t]
\centering
\includegraphics[width=\columnwidth]{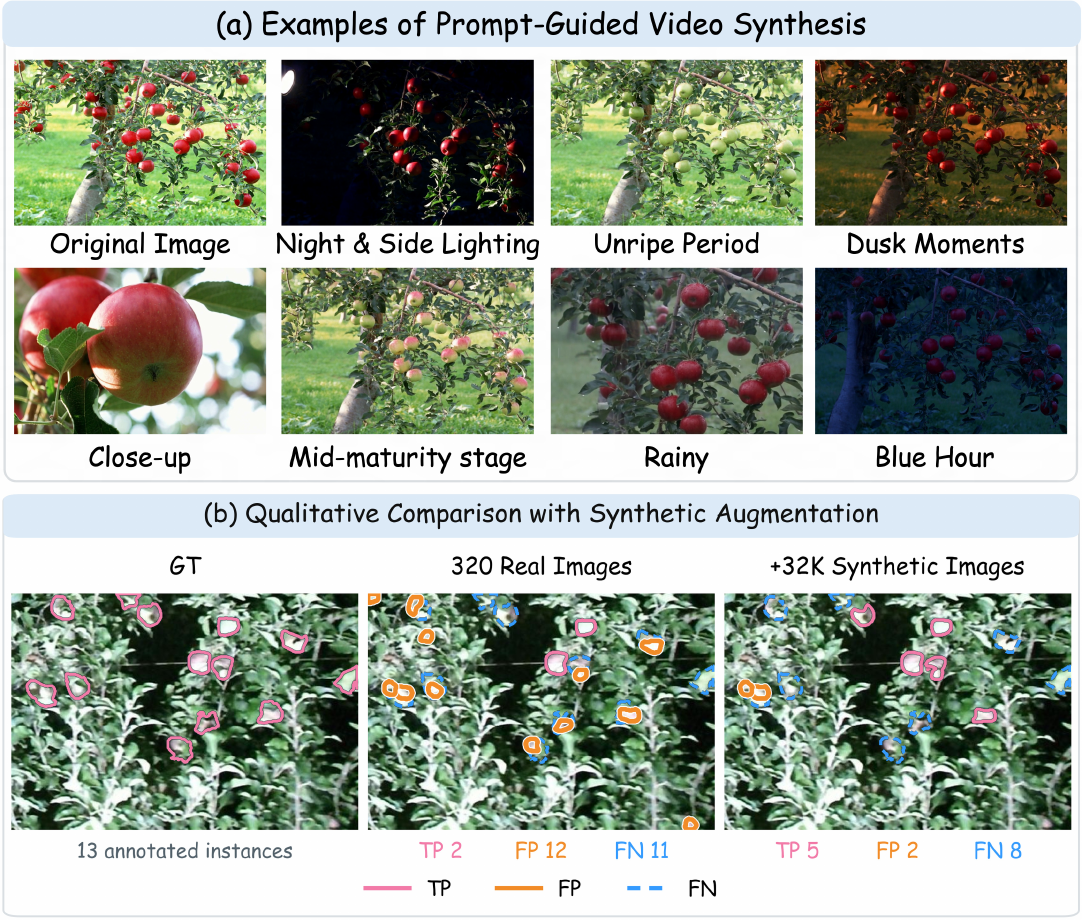}
\caption{Video augmentation and held-out KFuji predictions.
(a) Generated lighting, ripeness, and viewpoint variations.
(b) Predictions trained on 320 human-labeled images with or without synthetic frames (conf.~0.30, IoU~0.50).}
\label{fig:synthetic}
\vspace{-8pt}
\end{figure}
\setcounter{topnumber}{2}

\subsection{Synthetic Augmentation}
\label{sec:synthetic}
To test whether generative synthesis alleviates label scarcity, we introduce \textbf{BLOOM}, a prompt-guided video augmentation pipeline built on Seedance~2.5 (\Cref{fig:synthetic}a; details in supplement). From each seed image, we synthesize 20 five-second clips (100 frames at 1~fps) varying lighting, canopy density, viewpoint, and ripeness. Using 320 seeds from 16 PaperFruits subsets (20 each) yields 32,000 frames; retaining fruit-present frames adds 31,308 SAM~3-labeled frames to 320 human-labeled images, keeping all five test domains held out.

As shown in \Cref{tab:synthetic}, synthetic expansion delivers broad gains when real fruit supervision is limited, lifting average mask AP$_{50:95}$ by +4.85 from official weights (58.90 to 63.75) and by +1.04 under SA-1B \emph{Pre} initialization (62.62 to 63.65), while suppressing canopy false positives on held-out KFuji (\Cref{fig:synthetic}b; FP drops from 12 to 2 as TP rises from 2 to 5). Notably, KFuji and WebOOD benefit consistently across all three initializations, still gaining +2.47 and +0.68 AP$_{50:95}$ even after \emph{Pre}+\emph{Mid} (where overall average mask AP$_{50:95}$ dips by 2.22 once 570K+ real images have covered common viewpoint and ripeness variations). Unlike agricultural collections captured under routine illumination, nighttime KFuji and photography-rich WebOOD exhibit strong backlighting, warm golden/dusk casts, and deep-shadow contrast, precisely the lighting regimes targeted by our video synthesis. Generative augmentation thus acts as a versatile booster in scarce-data regimes and a targeted complement for underrepresented lighting conditions.

\subsection{Cross-Domain Recipe Validation}
\label{sec:cross-domain}
To verify that our \emph{Pre}--\emph{Mid}--\emph{Post} paradigm generalizes beyond agriculture, 
we apply the same recipe, with zero fruit data or fruit-trained weights, to underwater salient instances 
(USIS10K~\cite{lian2024diving}), 
microscopy cells (LIVECell~\cite{edlund2021livecell}), 
and occluded humans (CIHP~\cite{gong2018cihp} training, OCHuman~\cite{zhang2019pose2seg} testing) 
using RF-DETR-XL (37.79M parameters). In each task, \emph{Mid} replaces manual labeling with automated 
pseudo-masks (Qwen-3.8-max~\cite{qwen2026qwen38}+SAM~2.1~\cite{ravi2024sam2} for USIS10K and SAM~3~\cite{carion2026sam3} for the other tasks), 
while \emph{Post} uses only 1\% human labels on USIS10K/LIVECell (74/32 images) and 0.1\% on humans (28 images); 
selection-label budgets and protocols are detailed in the supplement.

As shown in \Cref{tab:cross-domain-transfer}, automated \emph{Mid} supervision is consistently the primary driver of label efficiency. Even where SA-1B \emph{Pre} yields mixed returns (helping USIS10K and LIVECell, yet slightly trailing on OCHuman where public COCO weights already encode human priors), adding \emph{Mid} alone lifts limited-label \emph{Post} by +10.00, +28.41, and +10.50 AP$_{50:95}$ on USIS10K, LIVECell, and OCHuman, respectively, and reaches 61.25 and 42.28 AP$_{50:95}$ on USIS10K and LIVECell with 0\% human labels (\emph{Pre}+\emph{Mid}). Paired with minimal \emph{Post} calibration, our compact 37.79M model recovers 93.4\%--95.8\% of its 100\%-GT upper bound across all benchmarks (64.18 vs.\ 68.74, 45.79 vs.\ 47.79, and 48.23 vs.\ 51.14 AP$_{50:95}$), while matching or surpassing larger 100\%-supervised author baselines on USIS10K (64.30, $\sim$701M) and OCHuman (34.70, 80.85M).

\begin{table}[!t]
\centering
\caption{Cross-domain validation of the SAF training recipe. Limited-label Post uses 1\% GT for USIS10K/LIVECell and 0.1\% for humans. All metrics are mask AP, following the authors' original evaluation settings. $\dagger$Author-reported results~\cite{lian2024diving,edlund2021livecell,kolomiiets2026sampose2seg}.}
\label{tab:cross-domain-transfer}
\label{tab:usis-transfer}
\label{tab:livecell-transfer}
\begingroup
\fontsize{8}{10}\selectfont
\setlength{\tabcolsep}{1.5pt}
\renewcommand{\arraystretch}{1.0}
\begin{tabularx}{\columnwidth}{@{}l*{6}{>{\centering\arraybackslash}X}@{}}
\toprule
\multirow{2}{*}{Model / route}
& \multicolumn{2}{c}{USIS10K} & \multicolumn{2}{c}{LIVECell} & \multicolumn{2}{c}{OCHuman} \\
\cmidrule(lr){2-3}\cmidrule(lr){4-5}\cmidrule(l){6-7}
 & AP$_{50:95}$ & AP$_{50}$ & AP$_{50:95}$ & AP$_{50}$ & AP$_{50:95}$ & AP$_{50}$ \\
\midrule
\multicolumn{7}{@{}l}{\color{tablegroupgray}\itshape Full supervision} \\[1pt]
Authors' model$^{\dagger}$ & 64.30 & 84.90 & \textbf{47.89} & \textemdash & 34.70 & \textemdash \\
Post & \textbf{68.74} & \textbf{87.51} & 47.79 & 80.45 & \textbf{51.14} & \textbf{64.95} \\
\midrule
\multicolumn{7}{@{}l}{\color{tablegroupgray}\itshape Limited human labels} \\[1pt]
Post & 52.31 & 68.89 & 16.61 & 33.29 & 37.73 & 50.26 \\
Pre + Post & 57.41 & 74.95 & 33.09 & 60.62 & 35.11 & 53.50 \\
Pre + Mid & 61.25 & 81.24 & 42.28 & 74.26 & 35.58 & 47.15 \\
Mid + Post & 62.31 & 81.74 & 45.02 & 78.46 & \textbf{48.23} & \textbf{62.79} \\
Pre + Mid + Post & \textbf{64.18} & \textbf{83.57} & \textbf{45.79} & \textbf{79.28} & 46.86 & 61.94 \\
\bottomrule
\end{tabularx}
\par\vspace{3pt}
\begin{minipage}{\columnwidth}
\fontsize{8}{9.5}\selectfont
\end{minipage}
\endgroup
\end{table}

\section{Conclusion}
\label{sec:discussion}
\label{sec:conclusion}
In this work, we demonstrated that achieving foundation-model-level generalization in specialized visual domains does not require deploying massive, prompt-dependent models at runtime. 
Through progressive \textit{Pre-Mid-Post} supervision across WebFruits and PaperFruits, \textbf{SegAnyFruit} (SAF) distills broad foundation priors into compact, prompt-free segmenters capable of real-time onboard inference.
Our findings highlight three practical takeaways for domain-specific perception: (1) learning a unified visual concept across diverse environments yields strong zero-shot transfer while serving as a resilient initialization that curbs catastrophic forgetting during local adaptation; (2) large-scale upstream supervision acts as a potent annotation multiplier, recovering near-full-data accuracy with as little as 1\% human-verified domain labels; and (3) this progressive distillation recipe extends naturally beyond agriculture to other data-scarce segmentation tasks such as underwater and cellular imagery.

\vspace{0.2em}
\textbf{Limitations and future work.}
Our current framework has three primary limitations. First, automatically generated pseudo-masks still retain errors in boundary quality and instance granularity. Second, although the benefits of generic mask pretraining vary across architectures and downstream tasks, we have not yet established when it is most effective. Third, SAF treats fruit as a unified visual concept and does not distinguish individual species. In future work, we will focus on improving automated annotation quality for sharper boundaries and consistent granularity, extending SAF to fine-grained species recognition, and conducting larger-scale studies across domains and architectures to clarify how training stages should be best combined.
\FloatBarrier

{\small
\bibliographystyle{ieeenat_fullname}
\bibliography{references}
}

\clearpage
\twocolumn[{\centering\Large\bfseries Supplementary Material\par\vspace{12pt}}]
\input{supplementary}

\end{document}

%% file: supplementary.tex
\raggedbottom
% Natural float flow: compact spacing without forced column/page breaks.
\setcounter{topnumber}{2}
\setcounter{bottomnumber}{2}
\setcounter{totalnumber}{4}
\setcounter{dbltopnumber}{2}
\renewcommand{\topfraction}{0.90}
\renewcommand{\bottomfraction}{0.85}
\renewcommand{\textfraction}{0.08}
\renewcommand{\floatpagefraction}{0.60}
\renewcommand{\dbltopfraction}{0.90}
\renewcommand{\dblfloatpagefraction}{0.60}
\setlength{\textfloatsep}{10pt plus 2pt minus 2pt}
\setlength{\dbltextfloatsep}{10pt plus 2pt minus 2pt}
\setlength{\floatsep}{6pt plus 1pt minus 1pt}
\appendix
\setcounter{figure}{0}
\setcounter{table}{0}
\renewcommand{\theHfigure}{supp.\arabic{figure}}
\renewcommand{\theHtable}{supp.\arabic{table}}
\renewcommand{\thefigure}{S\arabic{figure}}
\renewcommand{\thetable}{S\arabic{table}}
\paragraph{Overview.}
This supplement provides implementation details and additional analyses of
SegAnyFruit (SAF).
\Cref{sec:app-data} describes the fruit datasets, annotation workflows,
and mask curation.
\Cref{sec:supp-training} details the training configurations, evaluation
protocols, and edge deployment benchmarks.
\Cref{sec:app-complete} reports results across model variants and data
budgets on all seven test sets, along with zero-shot detection results on
MetaFruit.
\Cref{sec:app-qualitative} compares segmentation results across methods
and scenes.
\Cref{sec:app-generation} describes how BLOOM generates videos from
reference images for data augmentation.
\Cref{sec:app-cross-task} examines the transferability of the SAF framework
to underwater, cell, and human segmentation, reporting data budgets,
per-cell-type results, and qualitative comparisons.
The SAF code repository and video demos accompany this supplement,
with step-by-step instructions for reproducing the pipeline.

\section{Dataset Construction and Annotation}
\label{sec:app-data}

\subsection{WebFruits collection and curation}
\label{sec:app-curation}
\begin{figure}[!t]
\centering
\captionsetup{skip=4pt}
\includegraphics[width=\columnwidth]{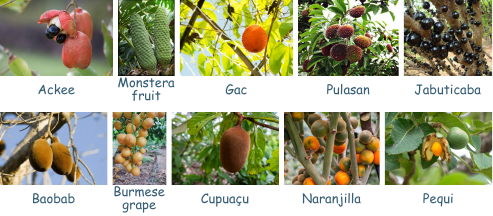}
\caption{\textbf{Fruit diversity in WebOOD.} Examples from
\mbox{\Cref{tab:app-ood-vocabulary}} illustrate varied fruit shapes, surface textures,
and growth.}
\label{fig:app-webood-gallery}
\end{figure}

We collect WebFruits from search engines, Flickr, iNaturalist, and GBIF.
Queries combine common and scientific fruit names in multiple languages
with growing environments, ripeness, and capture conditions. This covers
fruit on trees, vines, and bushes, as well as harvested fruit in markets
and homes. \Cref{tab:app-web-vocabulary} lists the 127 collection entries,
including named varieties. WebOOD uses 38 separate fruit names
(\Cref{tab:app-ood-vocabulary}) excluded from the WebFruits queries.
\Cref{fig:app-webood-gallery} shows ten representative WebOOD fruit types.

We use SAM~3 for its speed in batch annotation of the WebFruits training
pool. \Cref{fig:app-data-construction}(a) illustrates category-aware
curation: remove duplicate masks, keep individual bananas and whole
blackberries, and separate fruits in mixed scenes. Overlap and containment
rules handle duplicates and groups, while a VLM distinguishes whole
blackberries from drupelets in ambiguous cases. Together, these rules
keep annotation units consistent across fruit types, accounting for differences
in morphology and whether fruits appear individually or in groups.

WebTest is held out from WebFruits. For the initially unlabeled WebTest
and WebOOD images, we generate preliminary masks with Qwen-3.8-max and
SAM~2.1 (\Cref{sec:app-point-annotation}), then manually review and correct
them to obtain the final test annotations.

\subsection{PaperFruits sources and annotation routes}
PaperFruits comprises 36 source collections: 31 for training and
validation and five agricultural test sources.
\Cref{tab:app-paper-sources} lists the 31 training and validation
collections, their source annotations, annotation routes, and image and
mask counts after curation, highlighting differences in collection size
and available supervision.

\begin{figure*}[!t]
\centering
\captionsetup{skip=4pt}
\includegraphics[width=0.8\textwidth]{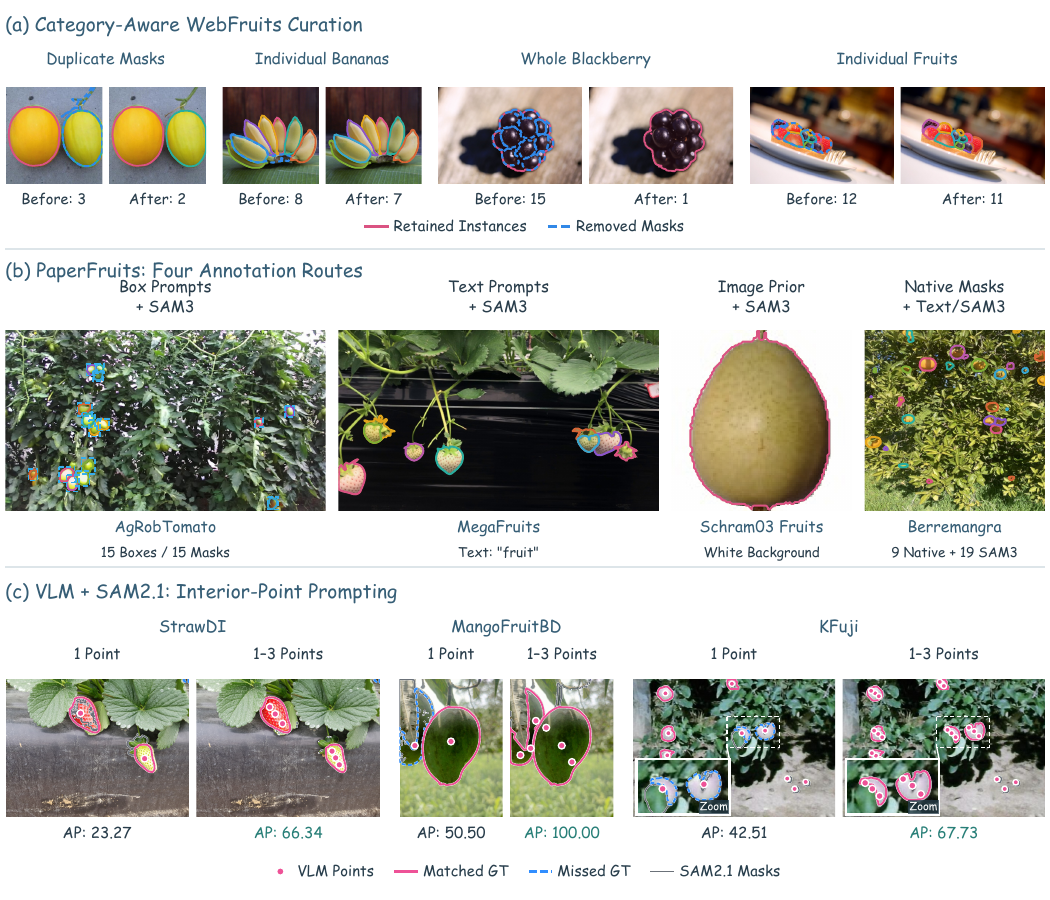}
\vspace{-3pt}
\caption{\textbf{Dataset construction and annotation.}
(a) WebFruits masks before and after curation.
(b) Four PaperFruits annotation routes, with cyan source boxes.
(c) Single-point and 1--3-point prompting. GT matching uses IoU $0.50$.
Mask AP$_{50:95}$ is computed per full image.}
\label{fig:app-data-construction}
\end{figure*}

For the training and validation collections, we select annotation routes
by manually reviewing each dataset's images and available annotations. We retain suitable native masks directly,
as in WGISD. For mask generation or supplementation,
\Cref{fig:app-data-construction}(b) shows four common routes: (1) \emph{box prompts}: we prompt SAM~3 with the source
bounding boxes in AgRobTomato; (2) \emph{text prompts}: we use fruit text
prompts with SAM~3 for MegaFruits; (3) \emph{image priors}: we estimate
foreground regions against Schram03's white backgrounds to generate box
and point prompts for SAM~3; and (4) \emph{native mask fusion}: we combine
Berremangra's original masks with text-prompted SAM~3 masks.
When two masks overlap by more than 90\% of each mask's area, we retain
the higher-scoring mask. We manually inspect a random 5\% of images.

We convert all sources to a common fruit-instance format, merging
source categories into one fruit class while preserving individual masks.
Each exported image retains its source collection, original path, and
annotation provenance, allowing the final labels to be traced back to
their source images and annotation routes.

The five agricultural test sources are MSU Apple~\cite{chu2023o2rnet},
KFuji~\cite{genemola2019kfuji}, MangoFruitBD~\cite{islam2026mangofruitbd},
StrawDI~\cite{perez2020strawdi}, and Rob2Pheno~\cite{afonso2020tomato}.
We exclude their training and validation splits from PaperFruits
training.

\subsection{Efficient automatic annotation}
\label{sec:app-point-annotation}

\begin{table}[!t]
\centering\footnotesize
\caption{\textbf{Prompt refinement and mask aggregation for VLM + SAM2.1.}
Average Mask AP$_{50:95}$ over seven fruit datasets.}
\label{tab:app-annotation-strategies}
\vspace{-6pt}
\begingroup
\setlength{\tabcolsep}{2pt}
\renewcommand{\arraystretch}{1.18}
\begin{tabular*}{\columnwidth}{@{\extracolsep{\fill}}lcccrc@{}}
\toprule
\makecell[l]{Prompt\\text} & \makecell{Points\\per fruit} &
\makecell{Candidates\\kept} & Dedup. &
\makecell{Mask\\AP$_{50:95}$} & \makecell{VLM\\calls} \\
\midrule
Original & 1 & Top-1 & No & 58.63 & 1 \\
Original & 1 & Top-1 & Yes & 59.35 & 1 \\
Refined & 1 & Top-1 & Yes & 60.21 & 1 \\
Refined & 1--3 & Top-1 & Yes & 60.41 & 1 \\
\midrule
\makecell[l]{Original +\\Refined} & 1 / 1--3 & Top-1 / Top-1 & Yes & \textbf{62.49} & 2 \\
\bottomrule
\end{tabular*}
\par\vspace{2pt}
\begin{minipage}{\columnwidth}\scriptsize
\textit{Original} uses one point. \textit{Refined} uses separated
interior points, with only the first used in the 1-point setting.
Top-1 selects the highest-scoring of three SAM~2.1 masks per point group.
The last row merges masks from both prompts before duplicate suppression
(Dedup.). Calls are VLM requests per image.
\end{minipage}
\endgroup
\end{table}

% Keep the annotation-strategy table with A.3 before the wide inventories.
% Tables S2/S3 share a wide float; the compact gallery stays with the data section.
% Compact row/rule spacing; retain the original vocabulary font sizes.
\begin{table*}[!t]
\centering
\begingroup
\centering
\captionsetup{position=top,skip=9pt} % Table S2: caption-to-table gap
\fontsize{7}{8.3}\selectfont
\setlength{\tabcolsep}{2pt}
\setlength{\aboverulesep}{0.8pt}
\setlength{\belowrulesep}{1.2pt}
\setlength{\lightrulewidth}{0.25pt}
\renewcommand{\arraystretch}{1.05}
\caption{WebFruits collection vocabulary: 127 fruit names and varieties.}
\label{tab:app-web-vocabulary}
\vspace{-6pt}
\begin{tabularx}{\textwidth}{@{}*{12}{>{\centering\arraybackslash}X}@{}}
\toprule
Blueberry & Mulberry & Cherry & Green Grape & Purple Grape & Apple & Pear & Loquat & Orange & Mandarin & Dekopon (Shiranui) & Baba Orange \\
\midrule
Sugar Tangerine & Lemon & Peach & Plum & Banana & Mango & Lychee & Longan & Papaya & Passion Fruit & Pineapple & Kiwifruit \\
\midrule
Watermelon & Cherry Tomato & Tomato & Pomegranate & Persimmon & Star Fruit & Fig & Chinese Crabapple & Plumleaf Crabapple & Chinese White Pear & Asian Pear & Ussurian Pear \\
\midrule
European Pear & Chinese Hawthorn & Quince Fruit & Yellow Peach & Nectarine & Flat Peach & European Plum & Apricot & Chinese Cherry & Sour Cherry & Japanese Apricot Fruit & Chinese Jujube Fruit \\
\midrule
Sour Jujube Fruit & Date Palm Fruit & Chinese Olive Fruit & Olive Fruit & Garden Strawberry & Lowbush Blueberry & American Cranberry & Bog Bilberry & Red Raspberry & Blackberry Fruit & Gooseberry Fruit & Blackcurrant Fruit \\
\midrule
Redcurrant Fruit & Black Mulberry Fruit & Golden Kiwifruit & Green Kiwifruit & Hardy Kiwi Fruit & Black Sapote Fruit & Goji Berry & Sea Buckthorn Berry & Honeyberry Fruit & European Elderberry & Wild Strawberry & Black Chokeberry Fruit \\
\midrule
Chestnut Rose Fruit & Schisandra Berry & Akebia Fruit & Mock Strawberry Fruit & Blood Orange & Clementine Fruit & Pomelo Fruit & Grapefruit & Key Lime Fruit & Finger Lime Fruit & Kumquat Fruit & Citron Fruit \\
\midrule
Buddha's Hand Citron & Bitter Orange Fruit & Sweet Lime Fruit & Trifoliate Orange Fruit & Yuzu Fruit & W Murcott Mandarin & Shatian Pomelo & Daidai Orange Fruit & Muskmelon Fruit & Hami Melon & Bailan Melon & Netted Melon \\
\midrule
Horn Honey Melon & African Horned Melon & Pepino Melon Fruit & Snake Gourd Fruit & Plantain Banana Fruit & Dragon Fruit & Yellow Dragon Fruit & Rambutan Fruit & Durian Fruit & Mangosteen Fruit & Guava Fruit & Strawberry Guava \\
\midrule
Sugar Apple Fruit & Soursop Fruit & Custard Apple Fruit & Cherimoya Fruit & Wax Apple Fruit & Rose Apple Fruit & Avocado Fruit & Yellow Passion Fruit & Jackfruit & Breadfruit & Cempedak Fruit & Bilimbi Fruit \\
\midrule
Sapodilla Fruit & Canistel Fruit & Abiu Fruit & Mamey Sapote Fruit & Langsat Fruit & Snake Fruit & Tamarind Fruit &  &  &  &  &  \\
\bottomrule
\end{tabularx}
\endgroup
\par\vspace{10pt}
\begingroup
\centering
\captionsetup{position=top,skip=3pt} % Table S3: caption-to-table gap
\footnotesize
\setlength{\tabcolsep}{2pt}
\setlength{\aboverulesep}{0.8pt}
\setlength{\belowrulesep}{1.2pt}
\setlength{\lightrulewidth}{0.25pt}
\renewcommand{\arraystretch}{1.08}
\caption{WebOOD collection vocabulary: 38 additional fruit types.}
\label{tab:app-ood-vocabulary}
\begin{tabularx}{\textwidth}{@{}*{6}{>{\centering\arraybackslash}X}@{}}
\toprule
Feijoa & Jabuticaba & Lucuma & Cupuaçu & Ackee & Ambarella \\
\midrule
African star apple & Baobab fruit & Miracle fruit & Gac fruit & Santol fruit & Pulasan \\
\midrule
Marang fruit & Burmese grape & Velvet apple & Bignay & Governor's plum & Imbe fruit \\
\midrule
Kei apple & Monkey orange & Safou fruit & Mamoncillo & Pitomba & Star apple \\
\midrule
Yellow mombin & Red mombin & Velvet tamarind & Monstera fruit & Chupa-chupa & Batuan fruit \\
\midrule
Achacha & Mangaba fruit & Pequi fruit & Bacuri fruit & Sawo kecik & Rukam fruit \\
\midrule
Marula fruit & Naranjilla &  &  &  &  \\
\bottomrule
\end{tabularx}
\endgroup
\end{table*}

% Two-row single-column Figure S2 preserves all ten full-frame examples.

% Inlined from data/appendix_dataset_20261119/paperfruits_sources.tex
% Allow this source inventory at the page top so it does not block later tables.
\begin{table*}[!t]
\centering\footnotesize
\setlength{\tabcolsep}{3pt}
\renewcommand{\arraystretch}{1.3}
\caption{PaperFruits training and validation sources. Original annotations describe the source data; our route describes the masks retained for training. Images and masks are counted after curation across the training and validation splits.}
\label{tab:app-paper-sources}
\vspace{-3pt}
\begin{tabularx}{\textwidth}{@{}>{\centering\arraybackslash}p{.09\textwidth}>{\raggedright\arraybackslash}p{.16\textwidth}>{\raggedright\arraybackslash}X>{\raggedright\arraybackslash}p{.16\textwidth}>{\raggedright\arraybackslash}p{.17\textwidth}rr@{}}
\toprule
\makecell{Fruit\\category} & Collection & Paper / source & Original annotations & Our annotation route & Images & Masks \\
\midrule
\multirow{7}{=}{\centering Apple} & AppleGrowthVision & AppleGrowthVision~\cite{app_applegrowth} & Boxes; growth stage & Text + SAM3 & 7,092 & 547,656 \\
 & Elstar & AmodalAppleSize~\cite{app_amodaldata} & Size measurements & Text + SAM3 & 1,070 & 9,975 \\
 & FujiSfM & Fuji-SfM~\cite{app_fujisfm} & Instance polygons & Text + SAM3 & 779 & 36,676 \\
 & PApple RGB-D & Amodal fruit sizing~\cite{app_amodalsizing} & Modal/amodal masks & Text + SAM3 & 2,895 & 15,222 \\
 & WSU Apple & WSU Apple~\cite{bhusal2019apple} & Boxes & Boxes/text + SAM3 & 2,172 & 68,183 \\
 & California orchard & In-house dataset & --- & Text + SAM3 & 1,292 & 82,443 \\
 & MinneApple & MinneApple~\cite{hani2020minneapple} & Instance polygons & Text + SAM3 & 670 & 35,847 \\
\midrule
\multirow{1}{=}{\centering Grape} & WGISD & WGISD~\cite{santos2020wgisd} & Masks; boxes & Native polygons & 242 & 3,579 \\
\midrule
\multirow{2}{=}{\centering Litchi} & Litchi-UAV & Litchi-SORT~\cite{app_litchisort} & Boxes & Boxes + SAM3 & 1,765 & 38,429 \\
 & UAVLitchi & UAVLitchi~\cite{app_uavlitchi} & Boxes (VOC) & Boxes/text + SAM3 & 3,742 & 32,005 \\
\midrule
\multirow{9}{=}{\centering Mixed fruits} & FruitBench & FruitBench~\cite{tjietfruitbench} & Class; growth stage & Text + SAM3 & 2,947 & 34,639 \\
 & FruitVision & FruitVision~\cite{bijoy2025fruitvision} & Quality classes & Text + SAM3 & 8,930 & 8,946 \\
 & MegaFruits & MegaFruits~\cite{wang2026sdmd} & Instance polygons & Text + SAM3 & 48,005 & 718,045 \\
 & Schram03 fruits & Schram03/fruits~\cite{schram03fruits} & Classes & Image prior + SAM3 & 19,661 & 19,661 \\
 & ToxiFruit & ToxiFruit~\cite{app_toxifruit} & Classes & Text + SAM3 & 5,954 & 5,954 \\
 & ACFR multifruit & Deep Fruit Detection~\cite{bargoti2017} & Boxes; semantic masks & Boxes/text + SAM3 & 2,568 & 16,420 \\
 & Fresh/rotten fruit & Fresh/rotten fruit~\cite{app_freshrotten} & Quality classes & Text + SAM3 & 3,025 & 12,404 \\
 & Orchard SLAM & 4D orchard mapping~\cite{app_4dorchard} & RGB sequences; 3D GT & Text + SAM3 & 13,115 & 269,071 \\
 & VegFru (Fru92) & VegFru~\cite{hou2017vegfru} & Classes & Text + SAM3 & 8,901 & 140,469 \\
\midrule
\multirow{3}{=}{\centering Citrus} & Berremangra & Berremangra~\cite{beldek2026berremangra} & Instance polygons & Native + text/SAM3 & 553 & 28,592 \\
 & CitDet & CitDet~\cite{app_citdet} & Boxes; raw images & Boxes/text + SAM3 & 1,049 & 74,983 \\
 & Oranges Field & Oranges in the field~\cite{orange2026} & Boxes (COCO) & Boxes + SAM3 & 4,714 & 39,765 \\
\midrule
\multirow{6}{=}{\centering Strawberry} & Dyson RGB & In-house dataset~\cite{tafuro2022strawberry} & Masks; keypoints & Native + text/SAM3 & 1,422 & 2,372 \\
 & Fruit-SLAM & In-house dataset & --- & Text + SAM3 & 3,991 & 161,150 \\
 & Strawberry samples & In-house dataset & Boxes (VOC) & Boxes + SAM3 & 1,813 & 5,207 \\
 & SPI Oak Shanjie & In-house dataset & --- & Text + SAM3 & 4,562 & 74,730 \\
 & SPI Oak Shunba & In-house dataset & --- & Text + SAM3 & 3,413 & 103,145 \\
 & Zwang COLMAP & In-house dataset & --- & Text + SAM3 & 2,282 & 114,071 \\
\midrule
\multirow{3}{=}{\centering Tomato} & AgRobTomato & AgRobTomato~\cite{app_agrobtomato} & Boxes (VOC) & Boxes + SAM3 & 395 & 5,190 \\
 & LaboroTomato & LaboroTomato~\cite{laborotomato} & Instance polygons & Text + SAM3 & 643 & 11,741 \\
 & Tomato Plant Factory & Tomato Plant Factory~\cite{app_tomatoplant} & Boxes (YOLO) & Boxes + SAM3 & 494 & 7,646 \\
\midrule
\multicolumn{5}{l}{Total (31 collections)} & 160,156 & 2,724,216 \\
\bottomrule
\end{tabularx}
\par\vspace{3pt}
\begin{minipage}{\textwidth}\scriptsize
Native denotes the source instance polygons. Text denotes fruit-name prompts; image prior denotes a foreground-derived box and point prompts.
\end{minipage}
\end{table*}

We combine task-aware localization by Qwen-3.8-max with point-guided
segmentation by SAM~2.1. The VLM interprets the annotation instructions,
identifies individual targets, and provides interior points for SAM~2.1
to delineate them. This interface supports both whole-fruit annotation
and contextual tasks such as underwater salient-instance segmentation,
where target selection depends on the scene.

All seven fruit test sets use human-annotated ground truth. For WebTest
and WebOOD, which initially lack annotations, the Qwen + SAM~2.1 pipeline
provides preliminary masks for subsequent manual review and correction.
We also use this pipeline to generate underwater training pseudo-labels.
\Cref{tab:app-annotation-strategies} compares prompting and mask-aggregation
strategies against the human annotations using cached outputs from all
seven fruit test sets.

First, the refined prompt asks the VLM to localize each whole fruit
and distribute up to three points across smooth visible skin, avoiding
boundaries, branches, and occluding leaves. The VLM also returns a
tight box; SAM~2.1 receives only the points. These cues improve mask
completeness
(\Cref{fig:app-data-construction}(c)). With identical mask processing,
prompt refinement raises average Mask AP$_{50:95}$ from 59.35 to
60.21 using the first point and to 60.41 using all points.
Second, for mask selection, we retain the highest-scoring of three
SAM~2.1 candidates per point group and suppress lower-scoring duplicates
whose intersection exceeds 90\% of both mask areas. This raises the
original prompt's average Mask AP$_{50:95}$ from 58.63 to 59.35.
Merging and deduplicating both prompts' masks yields
62.49 average Mask AP$_{50:95}$ (+3.86 over the baseline)
with two VLM calls per image.

% Allow the training text to fill the space below the dataset figure.

\section{Training and Evaluation}
\label{sec:supp-training}

\subsection{Experimental setup}
\label{sec:supp-training-setup}

\paragraph{Training.}
Mixed-precision training uses eight NVIDIA A800 GPUs.
Each model family sees approximately 1.5M SA-1B images during
\textit{Pre}, which ends when validation Mask AP$_{50:95}$ plateaus.
\textit{Pre}, \textit{Mid}, and \textit{Post} are validated on 5,000 SA-1B, 1,490 WebFruits,
and 10,070 PaperFruits images, respectively.
We select stage checkpoints by validation Mask AP$_{50:95}$
and start the next stage from its weights with a fresh optimizer
and learning-rate schedule.

\textbf{Model configurations.}
We use existing YOLO26-seg and RF-DETR architectures.
For DINOv3+Mask2Former, Pretraining jointly trains the backbone,
adapter, and segmentation head. \textit{Mid} and \textit{Post} freeze the backbone
and update only the adapter and head.

\textbf{Evaluation and adaptation.}
We report COCO Mask AP$_{50:95}$ and Mask AP$_{50}$, weighting
the seven fruit test sets equally.
For adaptation, 1-shot runs train for 30 epochs without validation.
The 5- and 10-shot settings each use one validation image, and
full-shot runs follow the original train/validation ratio of each dataset.

\subsection{Edge deployment and inference timing}
\label{sec:app-edge}

\paragraph{TensorRT throughput.}
We benchmark static batch-one engines on a Jetson Orin NX 8GB
with JetPack 7.2.1 and TensorRT 10.16.2, with the GPU fixed at
1173\,MHz. We use \texttt{trtexec} with a 500\,ms warmup and
200 iterations with spin-wait mode enabled. Engine FPS excludes data
transfers, host preprocessing, and mask postprocessing. FP32 permits
TensorRT's default TF32 tactics.

\textbf{Precision validation.}
On 150 validation images from 30 sources, FP16 changes Mask
AP$_{50:95}$ by $+0.05$, $-0.01$, and $-1.41$ points relative
to FP32 for RF-DETR-M, RF-DETR-L, and DINOv3+M2F, respectively.

\section{Complete Fruit Segmentation Results}
\label{sec:app-complete}
\label{sec:supp-data}

We report all model variants, domain-specific training
studies, and zero-shot detection on MetaFruit. The segmentation splits
match the main paper; \textit{Pre}, \textit{Mid}, and \textit{Post} use SA-1B, WebFruits, and
PaperFruits, respectively.

\subsection{Complete results across model configurations}
\label{sec:app-model-configurations}
\begin{figure}[t]
\centering
\captionsetup{skip=4pt}
\includegraphics[width=\columnwidth]{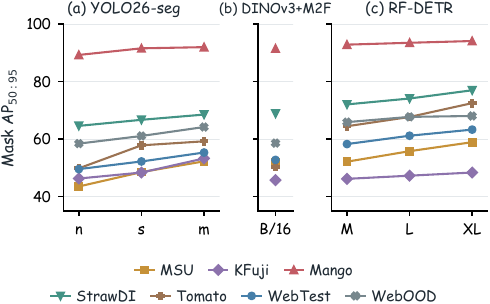}
\caption{Model configurations across seven test domains, all trained with \textit{Mid} + \textit{Post}.}
\label{fig:modelscale}
\end{figure}

\Cref{tab:app-model-complete} reports YOLO26n/s/m-seg and
RF-DETR-M/L/XL under \textit{Mid} + \textit{Post}, alongside main-paper
DINOv3+M2F results at two resolutions under \textit{Pre} + \textit{Mid} + \textit{Post}.
\Cref{fig:modelscale} compares seven \textit{Mid} + \textit{Post} variants, with
DINOv3+M2F using a 640-pixel short edge and a 1,024-pixel
long-edge cap. RF-DETR inputs span 432-624 pixels.

Scaling improves all seven domains in both families.
RF-DETR-M to XL gains 8.04 Mask AP$_{50:95}$ on Tomato, 6.79 on MSU,
and 1.26 on Mango (92.86 for M).
YOLO26n to m gains 8.74 on MSU and 9.40 on Tomato.

\begin{table*}[!t]
\centering
\fontsize{7}{8.5}\selectfont
\setlength{\tabcolsep}{1.2pt}\renewcommand{\arraystretch}{1.10}
\caption{Model configurations across seven test sets. Entries report
Mask AP$_{50:95}$/AP$_{50}$. The last column averages AP$_{50:95}$
equally across test sets. Bold marks the best scores.}
\label{tab:app-model-complete}
\vspace{-3pt}
\begin{tabular*}{\textwidth}{@{\extracolsep{\fill}}lccccccccr@{}}
\toprule
Model & Input & MSU & KFuji & Mango & StrawDI & Tomato & WebTest & WebOOD & Avg. AP$_{50:95}$ \\
\midrule
YOLO26n-seg & $640^2$ & 43.52\,/\,68.63 & 46.28\,/\,72.40 & 89.28\,/\,95.30 & 64.60\,/\,81.71 & 49.83\,/\,74.11 & 49.56\,/\,70.98 & 58.43\,/\,76.97 & 57.36 \\
YOLO26s-seg & $640^2$ & 48.51\,/\,73.21 & 48.37\,/\,72.93 & 91.58\,/\,96.80 & 66.71\,/\,84.23 & 57.83\,/\,79.95 & 52.22\,/\,72.60 & 61.08\,/\,77.84 & 60.90 \\
YOLO26m-seg & $640^2$ & 52.26\,/\,78.16 & \textbf{53.27}\,/\,\textbf{76.95} & 91.99\,/\,96.60 & 68.53\,/\,86.13 & 59.23\,/\,83.16 & 55.32\,/\,76.09 & 64.20\,/\,81.02 & 63.54 \\
\midrule
DINOv3+M2F & $640^2$ & 52.20\,/\,74.97 & 46.20\,/\,71.88 & 92.34\,/\,96.26 & 68.25\,/\,81.91 & 56.32\,/\,73.93 & 55.19\,/\,71.39 & 60.87\,/\,77.40 & 61.62 \\
DINOv3+M2F & $\ell{\leq}1024$ & 56.57\,/\,76.80 & 46.64\,/\,71.72 & 92.69\,/\,96.43 & 71.34\,/\,84.25 & 65.79\,/\,80.93 & 58.86\,/\,74.57 & 63.32\,/\,79.24 & 65.03 \\
\midrule
RF-DETR-M & $432^2$ & 52.14\,/\,77.19 & 46.17\,/\,73.46 & 92.86\,/\,96.62 & 72.07\,/\,87.89 & 64.47\,/\,86.45 & 58.27\,/\,77.82 & 65.90\,/\,81.98 & 64.55 \\
RF-DETR-L & $504^2$ & 55.79\,/\,79.60 & 47.27\,/\,74.31 & 93.51\,/\,96.82 & 74.12\,/\,88.60 & 67.70\,/\,\textbf{86.99} & 61.17\,/\,\textbf{78.96} & 67.73\,/\,\textbf{82.58} & 66.76 \\
RF-DETR-XL & $624^2$ & \textbf{58.93}\,/\,\textbf{80.53} & 48.37\,/\,75.49 & \textbf{94.12}\,/\,\textbf{97.11} & \textbf{76.99}\,/\,\textbf{90.81} & \textbf{72.51}\,/\,86.88 & \textbf{63.30}\,/\,78.88 & \textbf{68.04}\,/\,82.24 & \textbf{68.89} \\
\bottomrule
\end{tabular*}
\end{table*}

\subsection{Training stages and data budgets}
\label{sec:app-data-budgets}

\paragraph{Complementary supervision across domains.}
\Cref{tab:supp-sources} reports six training routes across seven test sets.
For DINOv3+M2F, adding \textit{Pre} to \textit{Mid} + \textit{Post} improves Mask AP$_{50:95}$
on all seven domains, including Tomato (50.31 to 65.79) and WebTest
(52.76 to 58.86).
For RF-DETR-XL, adding \textit{Post} to \textit{Mid} provides cultivated-scene supervision
from PaperFruits and raises Mask AP$_{50:95}$ on StrawDI (69.48 to 76.99)
and Tomato (64.82 to 72.51).

\textbf{Three data-budget settings.}
We scale WebFruits for \textit{Mid}, PaperFruits for \textit{Post}, or both by the same
fraction. The first and third settings compare official and \textit{Pre}
initializations. \textit{Post} also tests \textit{Pre} + full \textit{Mid} to assess broad fruit
supervision with limited PaperFruits labels. All curves use RF-DETR-XL
with zero-shot evaluation on the seven test sets, without target-dataset adaptation.
Pool fractions include the validation holdout. Each run retains its
checkpoints and training budget.

% Place Mid and joint scaling together before the Post study.
\begin{figure}[!t]
\centering
\captionsetup{skip=2pt}
\includegraphics[width=\columnwidth]{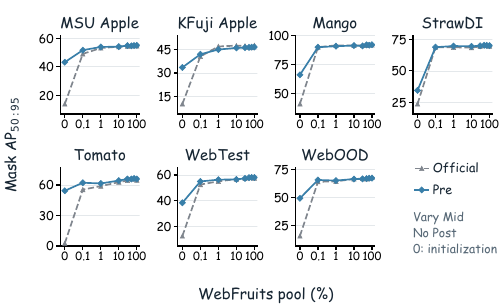}
\caption{Zero-shot segmentation as WebFruits \textit{Mid} data increases,
without \textit{Post}, from official weights or \textit{Pre}.}
\label{fig:app-mid-domains}
\end{figure}

\begin{figure}[!t]
\centering
\captionsetup{skip=2pt}
\includegraphics[width=\columnwidth]{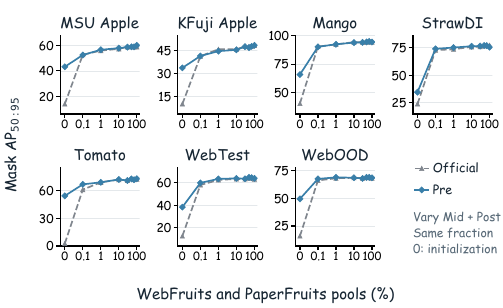}
\caption{Zero-shot segmentation as WebFruits \textit{Mid} and PaperFruits \textit{Post}
data increase together at equal fractions.}
\label{fig:app-joint-domains}
\end{figure}

\begin{table*}[!t]
\centering
\vspace{-3pt}
\captionsetup{position=top,skip=3pt}
\fontsize{7}{8.5}\selectfont
\setlength{\tabcolsep}{2.6pt}\renewcommand{\arraystretch}{0.95}
\caption{Training-stage ablation across seven test sets. Entries report
Mask AP$_{50:95}$/AP$_{50}$. Bold marks the best scores per model.}
\label{tab:supp-sources}
\begin{tabular*}{\textwidth}{@{\extracolsep{\fill}}cccccccccc@{}}
\toprule
Pre & Mid & Post & MSU & KFuji & Mango & StrawDI & Tomato & WebTest & WebOOD \\
\midrule
\multicolumn{10}{l}{\color{black!60}\itshape YOLO26m-seg} \\
$\times$ & $\checkmark$ & $\times$ & 39.20\,/\,73.50 & 42.51\,/\,72.17 & 87.05\,/\,95.45 & 53.49\,/\,80.39 & 36.49\,/\,66.79 & 38.33\,/\,62.46 & 50.06\,/\,67.74 \\
$\checkmark$ & $\checkmark$ & $\times$ & 39.95\,/\,74.44 & 41.14\,/\,70.59 & 87.46\,/\,96.08 & 51.29\,/\,78.17 & 41.12\,/\,75.89 & 40.48\,/\,66.47 & 51.83\,/\,70.47 \\
$\times$ & $\times$ & $\checkmark$ & 49.31\,/\,75.95 & 50.64\,/\,75.74 & \textbf{92.78}\,/\,\textbf{97.57} & \textbf{70.39}\,/\,\textbf{89.28} & 59.06\,/\,82.29 & 54.40\,/\,75.41 & 62.83\,/\,80.03 \\
$\checkmark$ & $\times$ & $\checkmark$ & 50.21\,/\,75.89 & 50.28\,/\,74.95 & 92.70\,/\,97.21 & 68.89\,/\,87.93 & 57.55\,/\,79.98 & 54.17\,/\,75.43 & 62.46\,/\,79.82 \\
$\times$ & $\checkmark$ & $\checkmark$ & \textbf{52.26}\,/\,\textbf{78.16} & \textbf{53.27}\,/\,\textbf{76.95} & 91.99\,/\,96.60 & 68.53\,/\,86.13 & \textbf{59.23}\,/\,\textbf{83.16} & \textbf{55.32}\,/\,\textbf{76.09} & \textbf{64.20}\,/\,\textbf{81.02} \\
$\checkmark$ & $\checkmark$ & $\checkmark$ & 47.63\,/\,73.86 & 48.16\,/\,71.65 & 92.26\,/\,96.72 & 69.40\,/\,88.63 & 56.90\,/\,79.49 & 53.42\,/\,74.50 & 61.64\,/\,78.76 \\
\midrule
\multicolumn{10}{l}{\color{black!60}\itshape DINOv3+M2F} \\
$\times$ & $\checkmark$ & $\times$ & 49.52\,/\,75.33 & \textbf{49.18}\,/\,\textbf{75.54} & 91.48\,/\,96.75 & 61.72\,/\,78.32 & 40.61\,/\,55.05 & 47.54\,/\,66.16 & 56.80\,/\,71.95 \\
$\checkmark$ & $\checkmark$ & $\times$ & 53.63\,/\,76.02 & 44.71\,/\,69.51 & 91.88\,/\,96.51 & 66.10\,/\,83.34 & 64.49\,/\,\textbf{81.91} & 55.36\,/\,72.20 & 62.86\,/\,76.51 \\
$\times$ & $\times$ & $\checkmark$ & 51.37\,/\,74.34 & 41.75\,/\,69.31 & 92.48\,/\,\textbf{97.17} & 67.78\,/\,83.58 & 53.19\,/\,70.31 & 52.49\,/\,71.13 & 57.90\,/\,75.82 \\
$\checkmark$ & $\times$ & $\checkmark$ & 54.09\,/\,74.48 & 44.76\,/\,69.65 & 91.67\,/\,96.32 & 71.07\,/\,\textbf{86.01} & 63.51\,/\,78.58 & 56.70\,/\,73.61 & 61.55\,/\,78.04 \\
$\times$ & $\checkmark$ & $\checkmark$ & 51.97\,/\,74.54 & 45.71\,/\,72.20 & 91.50\,/\,96.91 & 68.87\,/\,83.17 & 50.31\,/\,66.90 & 52.76\,/\,70.83 & 58.60\,/\,75.54 \\
$\checkmark$ & $\checkmark$ & $\checkmark$ & \textbf{56.57}\,/\,\textbf{76.80} & 46.64\,/\,71.72 & \textbf{92.69}\,/\,96.43 & \textbf{71.34}\,/\,84.25 & \textbf{65.79}\,/\,80.93 & \textbf{58.86}\,/\,\textbf{74.57} & \textbf{63.32}\,/\,\textbf{79.24} \\
\midrule
\multicolumn{10}{l}{\color{black!60}\itshape RF-DETR-XL} \\
$\times$ & $\checkmark$ & $\times$ & 54.60\,/\,80.37 & 47.96\,/\,75.36 & 91.76\,/\,97.80 & 69.48\,/\,86.73 & 64.82\,/\,85.70 & 57.48\,/\,76.36 & 67.98\,/\,80.63 \\
$\checkmark$ & $\checkmark$ & $\times$ & 55.07\,/\,80.46 & 46.87\,/\,74.70 & 91.93\,/\,\textbf{97.95} & 69.88\,/\,86.92 & 65.96\,/\,87.39 & 58.04\,/\,76.81 & 67.30\,/\,80.85 \\
$\times$ & $\times$ & $\checkmark$ & 58.06\,/\,79.76 & 47.57\,/\,73.97 & 93.67\,/\,96.86 & 75.73\,/\,89.81 & 71.69\,/\,\textbf{87.70} & 63.00\,/\,79.39 & 68.32\,/\,\textbf{82.68} \\
$\checkmark$ & $\times$ & $\checkmark$ & 58.71\,/\,80.05 & 48.19\,/\,75.06 & \textbf{94.24}\,/\,97.62 & 76.46\,/\,90.28 & 72.90\,/\,87.63 & 63.36\,/\,79.22 & 67.91\,/\,82.03 \\
$\times$ & $\checkmark$ & $\checkmark$ & 58.93\,/\,\textbf{80.53} & 48.37\,/\,\textbf{75.49} & 94.12\,/\,97.11 & \textbf{76.99}\,/\,\textbf{90.81} & 72.51\,/\,86.88 & 63.30\,/\,78.88 & 68.04\,/\,82.24 \\
$\checkmark$ & $\checkmark$ & $\checkmark$ & \textbf{60.14}\,/\,80.46 & \textbf{48.39}\,/\,\textbf{75.49} & 94.21\,/\,97.06 & 75.33\,/\,90.03 & \textbf{72.95}\,/\,87.47 & \textbf{64.01}\,/\,\textbf{79.51} & \textbf{68.38}\,/\,82.66 \\
\bottomrule
\end{tabular*}
\end{table*}

\textbf{Scaling mid-training data.}
\Cref{fig:app-mid-domains} shows that useful fruit segmentation emerges
from a small WebFruits pool. At 0\%, SA-1B \textit{Pre} achieves
45.65 average Mask AP$_{50:95}$ across the seven tests, before any
training on WebFruits or PaperFruits. Its 28.53-point gain over official
initialization (17.12) provides a strong starting point for fruit segmentation.
\textit{Pre} improves the 0.1\% \textit{Mid} result on
six of seven tests, while the two initializations become closer as more
WebFruits data are available. The curves reveal the domain
response hidden by an overall average, including different saturation
rates across cultivated scenes and web imagery.

\textbf{Jointly scaling mid- and post-training data.}
When both pools grow together at the same sampling fraction
(\Cref{fig:app-joint-domains}), most of the improvement occurs between
0.1\% and 10\%. Later increases
yield smaller, domain-dependent changes. The \textit{Pre} initialization
provides the stronger low-budget starting point on five of seven
tests.

\begin{figure}[!t]
\centering
\captionsetup{skip=2pt}
\includegraphics[width=\columnwidth]{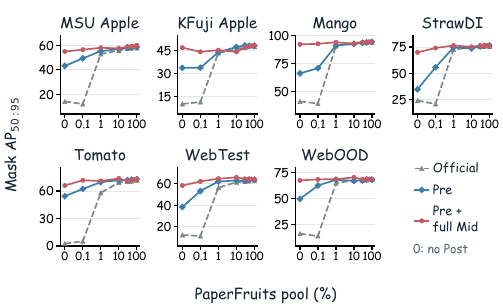}
\caption{Zero-shot segmentation as only PaperFruits \textit{Post} data increases,
from official weights, \textit{Pre}, or \textit{Pre} + full \textit{Mid}.}
\label{fig:app-post-domains}
\end{figure}

\begin{figure}[!t]
\centering
\captionsetup{skip=4pt}
\includegraphics[width=\columnwidth]{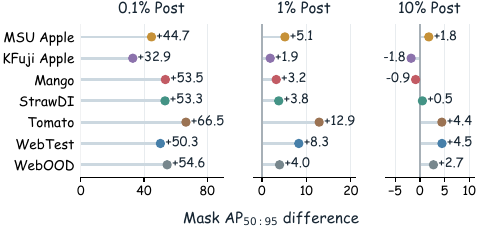}
\caption{Zero-shot Mask AP$_{50:95}$ difference: (\textit{Pre} + full \textit{Mid} + \textit{Post})
minus (Official + \textit{Post}), at identical PaperFruits fractions.}
\label{fig:app-low-label-gain}
\end{figure}

\textbf{Scaling post-training data.}
The benefit of a fruit-aware starting point is strongest when \textit{Post}
labels are scarce (\Cref{fig:app-post-domains,fig:app-low-label-gain}).
At 0.1\% \textit{Post}, \textit{Pre} + full \textit{Mid} improves over \textit{Pre} alone by 21.87 points
on Mango, 18.30 on StrawDI, and 10.39 on KFuji. At 1\% \textit{Post}, it
exceeds the official initialization on all seven tests. The gains
include 8.28 points on WebTest and 3.95 on WebOOD, showing that
limited refinement can retain the broad fruit coverage acquired
during \textit{Mid}.
The 0.1\% \textit{Post} pool contains 135 training and 15 validation images
sampled from PaperFruits, including only three training images from
tomato-specific sources.

% Keep the Post curves and their difference analysis together in one column.

\FloatBarrier

\begin{table}[!ht]
\centering
\captionsetup{position=top,skip=3pt}
\caption{Zero-shot Box AP$_{50:95}$ on MetaFruit.
SAF boxes are derived from instance masks on all 4,248 images,
without adaptation. Bold values exceed FMFruit-T~\cite{metafruit2025}.}
\label{tab:app-metafruit}
\fontsize{8}{9.5}\selectfont
\setlength{\tabcolsep}{2pt}
\renewcommand{\arraystretch}{1.10}
\begin{tabular*}{\columnwidth}{@{\extracolsep{\fill}}lccccc@{}}
\toprule
Model & Apple & Orange & Lemon & Grapefruit & Tangerine \\
\midrule
FMFruit-T      & 24.1           & 36.6           & 29.4           & 37.2           & 29.6           \\
\midrule
SAF-YOLO26m    & \textbf{40.65} & \textbf{38.07} & \textbf{40.76} & \textbf{39.04} & \textbf{36.02} \\
SAF-DINOv3+M2F & \textbf{37.61} & 36.58          & \textbf{38.06} & 35.01          & \textbf{31.06} \\
SAF-RF-DETR-XL & \textbf{42.37} & \textbf{40.54} & \textbf{42.53} & \textbf{39.67} & \textbf{37.17} \\
\bottomrule
\end{tabular*}
\end{table}

\setcounter{dbltopnumber}{3}
\renewcommand{\dbltopfraction}{0.9}

\begin{figure}[!t]
\centering
\captionsetup{skip=4pt}
\includegraphics[width=\columnwidth]{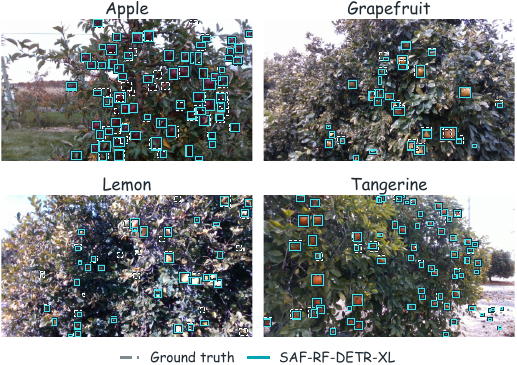}
\caption{Zero-shot SAF-RF-DETR-XL predictions on MetaFruit.
White dashed boxes show ground truth; cyan boxes are derived from
predicted instance masks with score $\geq0.30$.}
\label{fig:app-metafruit-qualitative}
\end{figure}

\subsection{Zero-shot detection from masks on MetaFruit}
\label{sec:app-metafruit}
SAF predicts instance masks on MetaFruit without adaptation.
Because MetaFruit~\cite{metafruit2025} provides box annotations for five
orchard fruits, we evaluate detection as a derived capability using
bounding boxes from these masks.
\Cref{tab:app-metafruit} compares SAF on all 4,248 available images
with the published FMFruit-T zero-shot results.
Box AP$_{50:95}$ uses a 100-detection limit; Box AP$_{50}$ and
Box AR$_{50:95}$ use 1,000.

Average Box AP$_{50:95}$ reaches 38.91 for YOLO26m, 35.66 for
DINOv3+M2F, and 40.46 for RF-DETR-XL. All exceed the published zero-shot
average of 31.38. YOLO26m and RF-DETR-XL also outperform the FMFruit-T
baseline in each of the five fruit categories. RF-DETR-XL gains 9.08
points over FMFruit-T on average, with the largest gain on Apple
(18.27 points).
\Cref{fig:app-metafruit-qualitative} shows transfer to dense, occluded
orchard scenes.

% END COMPLETE RESULTS 20261120

\FloatBarrier
% BEGIN QUALITATIVE SECTION D 20261120
\begin{figure*}[!t]
\captionsetup{hypcap=false,skip=4pt}
\centering
\includegraphics[width=\textwidth]{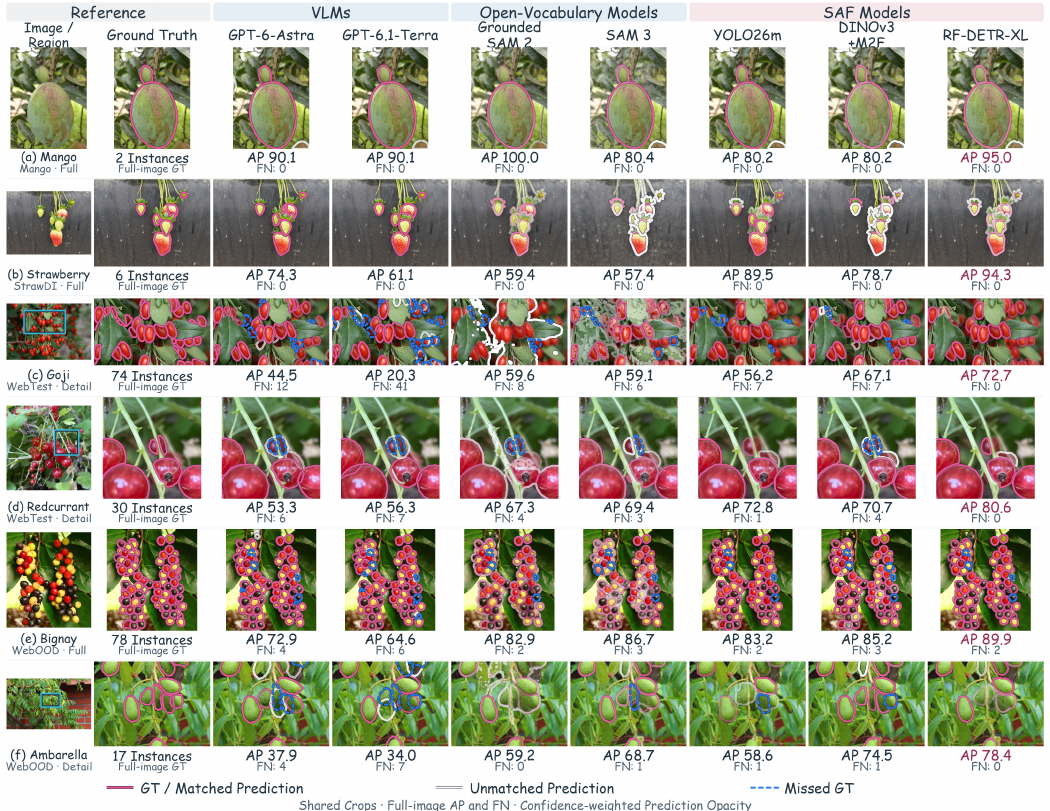}
\vspace{-3pt}
\caption{Qualitative fruit segmentation across scenes and fruit types.
Cyan boxes locate shared detail crops. Pink denotes GT contours and true positives (TP);
white, false positives (FP); dashed blue, missed GT (FN).
Predictions match GT one-to-one at mask IoU $\geq0.50$.
AP (Mask AP$_{50:95}$) and FN use full images.
All top-100 predictions are shown with confidence-weighted opacity.}
\label{fig:app-qualitative-methods}
\end{figure*}

\section{Qualitative Comparisons}
\label{sec:app-qualitative}
\Cref{fig:app-qualitative-methods} compares two VLMs, Grounded
SAM~2, SAM~3, and three SAF architectures. Full views show object
coverage, while shared crops reveal instance separation in dense
clusters. At IoU 0.50, RF-DETR-XL recovers all 74 annotated goji
berries and 30 redcurrants. SAM~3 misses six and three,
respectively. The crops show how RF-DETR-XL separates adjacent
fruits from one another and surrounding foliage.
Accurate fruit boundaries also matter. Both SAM~3 and RF-DETR-XL
match all six strawberries at IoU 0.50, with Mask AP$_{50:95}$ scores
of 57.4 and 94.3, respectively. This gap shows how stricter IoU
thresholds reveal differences in boundary quality. Bignay and
ambarella illustrate transfer to unfamiliar fruit types and dense
clusters, providing instance-level context for the aggregate scores.

\begin{figure*}[!t]
\centering
\includegraphics[width=\textwidth]{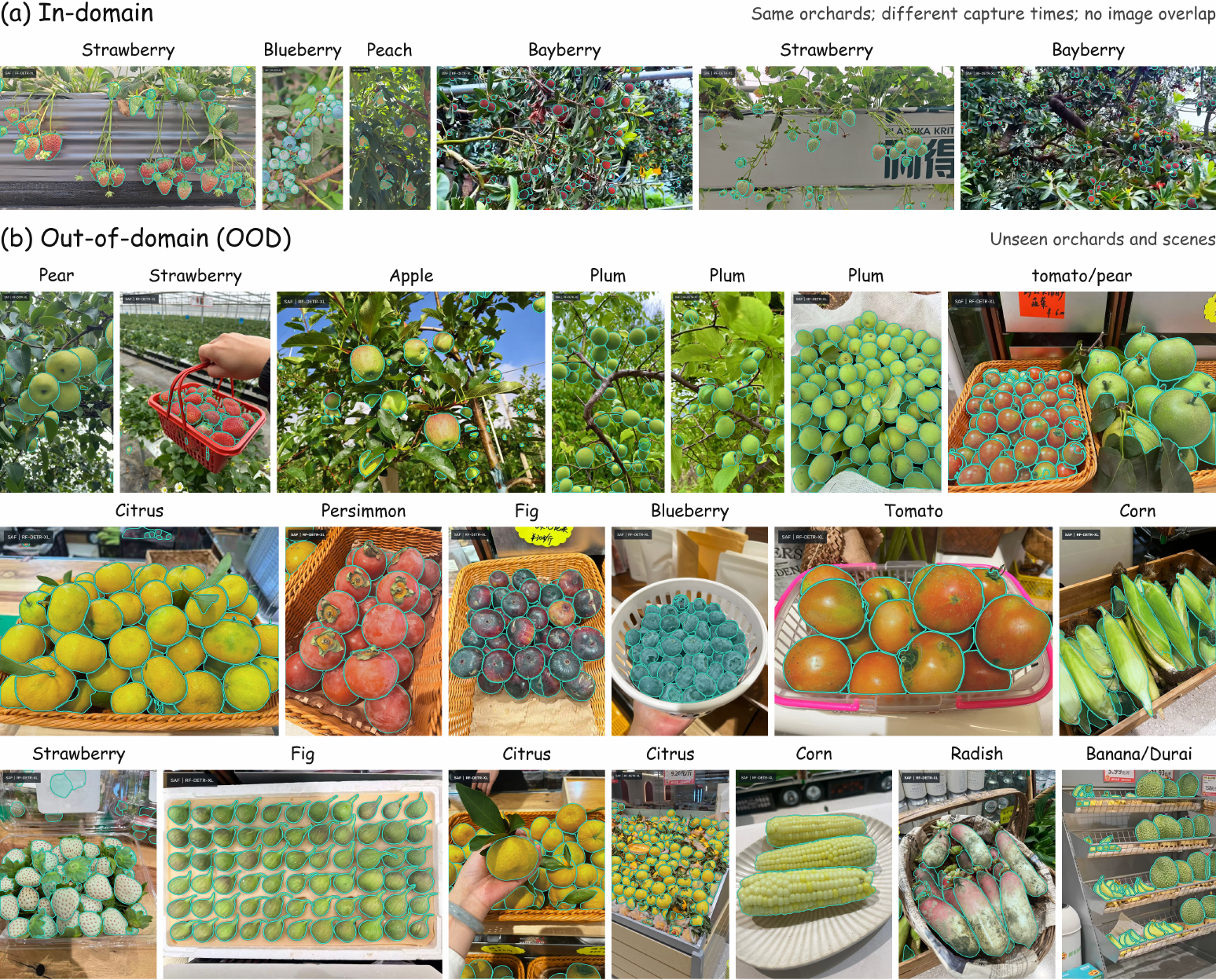}
\captionsetup{hypcap=false,skip=4pt}
\caption{\textbf{SAF-RF-DETR-XL predictions in familiar and unseen scenes.}
(a) In-domain orchard images from different capture times, without
overlap with training images.
(b) OOD examples from orchards and post-harvest scenes absent from
the training data. Cyan contours show predicted instance boundaries across
dense cultivation, harvesting, packaging, and retail settings.}
\label{fig:app-in-domain-ood}
\end{figure*}

\textbf{In-domain and out-of-domain scenes.}
\Cref{fig:app-in-domain-ood} shows SAF-RF-DETR-XL predictions across
cultivation and post-harvest settings. The first row shows in-domain
orchard images captured at different times from the training images,
with no image overlap. The remaining rows show out-of-domain (OOD)
orchards and post-harvest settings absent from training, including
harvest baskets, packed fruit, and retail displays.
In orchards, SAF separates fruit from dense foliage and neighboring
instances. In post-harvest scenes, it delineates individual fruits
in crowded arrangements. Together, these examples show how instance
separation transfers across backgrounds, viewpoints, and lighting
conditions. After large-scale training on diverse fruits, SAF also
delineates corn and radishes, illustrating transfer to some vegetable
categories.

% END QUALITATIVE SECTION D 20261120

% BEGIN VIDEO AUGMENTATION SECTION E 20261121
\section{BLOOM: Prompt-Guided Video Augmentation}
\label{sec:app-generation}

We introduce \textbf{BLOOM}, a prompt-guided video augmentation strategy
for fruit segmentation with limited annotations. From one reference
image, BLOOM generates videos under 20 controlled conditions and samples
their frames to expand the training set. Each prompt specifies
\emph{species morphology, a target condition, preservation constraints,
and a temporal schedule} to guide visual and temporal changes.

\begin{figure*}[!t]
\centering
\includegraphics[width=\textwidth]{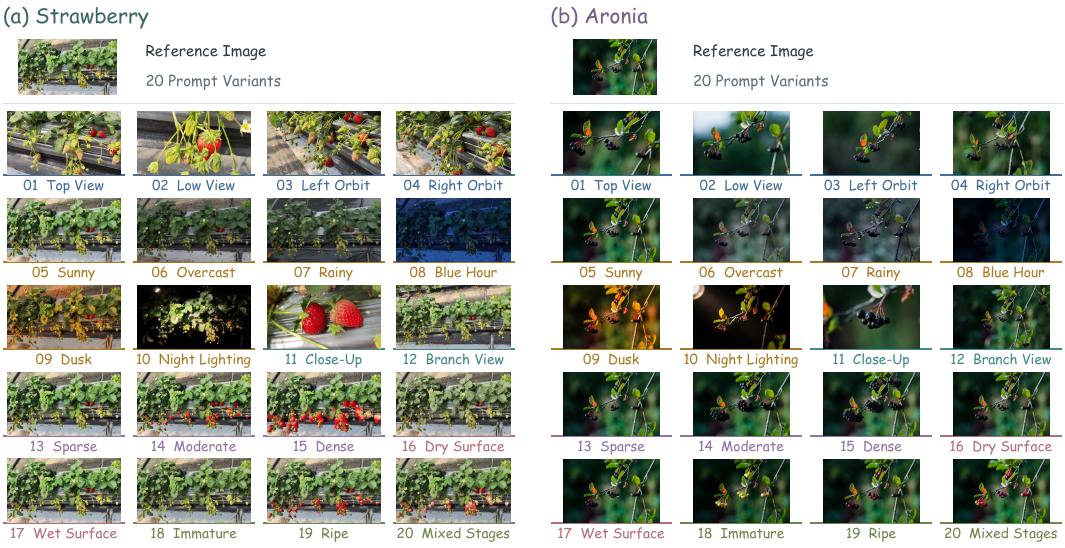}
\captionsetup{hypcap=false,skip=3pt}
\caption{\textbf{BLOOM: twenty variations from one reference image.}
Strawberry and Aronia each illustrate 20 conditions spanning viewpoint,
lighting, scale, density, surface appearance, and maturity.
References appear above the grids; matching numbers denote the same
requested condition, and label colors identify the six prompt families.}
\label{fig:app-video-prompts}
\end{figure*}

\textbf{A structured bank of 20 variations.}
Six complementary families cover four viewpoints (top, low, left orbit,
and right orbit), six lighting conditions (sunny, overcast, rainy, blue
hour, dusk, and night), two spatial scales (fruit close-up and branch
view), three densities (sparse, moderate, and dense), two surface states
(dry and wet), and three maturity settings (immature, ripe, and mixed).
These conditions address distinct sources of agricultural visual
variation: camera placement changes visible geometry and occlusion,
lighting changes appearance across weather and time of day, and density
changes the degree of crowding. Close-ups emphasize fruit texture and
shape; branch views retain foliage and cultivation context.

\textbf{Species-aware visual constraints.}
Morphological anchors describe the structures that identify each fruit.
For grapes, these include individual pedicels, a branching rachis, and
surface bloom; for Aronia, they include calyx scars, fine pedicels, and
serrated leaves. Lighting prompts specify light direction, shadows,
and reflections while preserving intrinsic skin color; maturity prompts
instead modify developmental appearance. The rainy condition explicitly
requests downward rain streaks, droplets on fruit and leaves, hanging
droplets at leaf tips, and a consistently wet background under diffuse
light. Preservation constraints maintain natural plant connections,
gravity, and occlusion by adjacent fruits, leaves, and branches.

\textbf{Temporal control.}
Prompts define a transition followed by a stable target condition.
A left orbit follows a $90^\circ$ arc around a stationary fruit cluster,
preserving continuous parallax before holding the final view.
Attribute transitions allow natural leaf and droplet motion after the
requested appearance stabilizes. Sampling along these trajectories
provides additional views from the same reference.

\Cref{fig:app-video-prompts} shows all 20 conditions for strawberry and
Aronia: branch views and fruit close-ups, daylight and nighttime scenes,
and contrasting fruit densities. The Aronia rainy example highlights
the weather transformation through visible rain streaks and leaf-tip droplets.
% END VIDEO AUGMENTATION SECTION E 20261121

% BEGIN CROSS-TASK APPENDIX 20261121
\section{Transferability of the SAF Framework}
\label{sec:app-cross-task}

\begin{figure*}[!t]
\centering
\includegraphics[width=\textwidth]{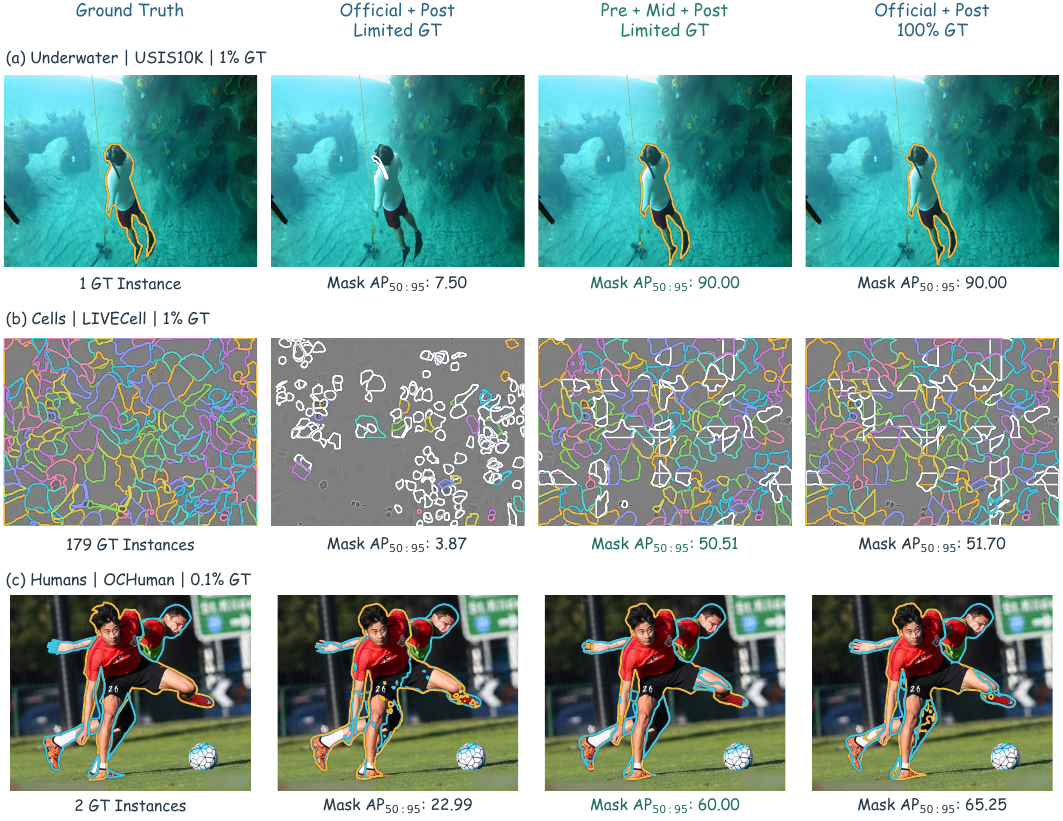}
\captionsetup{hypcap=false}\caption{\textbf{Qualitative transfer across three downstream tasks.}
Rows use 1\% \textit{Post} GT for underwater and cell segmentation and 0.1\%
for human segmentation. For visual comparison, each column shows the
$N$ highest-scoring masks, where $N$ is the GT instance count.
Matched instances share GT colors (mask IoU $\geq0.50$); unmatched
predictions are white.}
\label{fig:app-transfer-qualitative}
\end{figure*}

We complement the main-paper transfer results with data and annotation
details, per-cell-type results, and qualitative comparisons across the
three downstream tasks.

\begin{table}[!t]
\centering
\footnotesize
\setlength{\tabcolsep}{2pt}
\renewcommand{\arraystretch}{1.13}
\caption{\textbf{Data budgets for cross-task transfer.}
Counts refer to original images. \textit{Mid} lists automatically labeled training
images; \textit{Post} lists the limited-label GT subsets.}
\label{tab:app-transfer-data}
\begin{tabular*}{\columnwidth}{@{\extracolsep{\fill}}lrrlr@{}}
\toprule
Task & \makecell{Train\\pool} & Mid images & \makecell[l]{Mid\\annotation}
& \makecell{Post GT\\Images (fraction)} \\
\midrule
USIS10K & 7,442 & 7,350 & VLM + SAM\,2.1 & 74 (1\%) \\
LIVECell & 3,158 & 3,126 & SAM\,3 (\emph{cell}) & 32 (1\%) \\
\makecell[l]{CIHP $\rightarrow$\\OCHuman} & 28,104 & 27,728 & SAM\,3 (\emph{person}) & 28 (0.1\%) \\
\bottomrule
\end{tabular*}
\end{table}

\textbf{Data and annotation.}
\Cref{tab:app-transfer-data} summarizes automatic \textit{Mid} supervision and
human-labeled \textit{Post} data for underwater, cell, and human segmentation.
The \textit{Post} subsets contain 134
underwater instances, 12,137 cells, and 88 people, respectively.
USIS10K uses VLM-generated interior points to prompt SAM\,2.1;
LIVECell and CIHP use text-prompted SAM\,3 masks at confidence 0.5.
Images without generated labels are excluded from \textit{Mid} training.
LIVECell uses 3,158 unique training images after deduplication and
removal of validation overlaps. Human masks represent
whole people: training uses CIHP, and testing uses OCHuman.

\textbf{Training.}
All routes use RF-DETR-XL at 624-pixel input. \textit{Pre} initializes the model
from SA-1B; \textit{Mid} learns from the task's automatic masks, and \textit{Post} refines
it with the indicated GT subset. Each stage uses AdamW with learning
rates $5\times10^{-5}$ for the decoder and $5\times10^{-6}$ for the
encoder, weight decay $5\times10^{-4}$, and BF16 training.
Training runs for at most 100 epochs, stopping when both validation
Mask AP$_{50:95}$ and training loss remain unimproved for five epochs;
the validation-best EMA weights are evaluated.

\begin{table}[!t]
\centering
\small
\setlength{\tabcolsep}{3pt}
\renewcommand{\arraystretch}{1.08}
\caption{\textbf{Transfer across cell types.} Test Mask AP$_{50:95}$.
The first three routes use 32 \textit{Post} GT images; Full \textit{Post} uses 3,158.
\textit{Post} starts from official weights, and \textit{Pre} denotes SA-1B initialization.
Blue shading highlights \textit{Pre} + \textit{Mid} + \textit{Post}.}
\label{tab:livecell-types}
\begin{tabular*}{\columnwidth}{@{\extracolsep{\fill}}lrr>{\columncolor{cvprblue!12}}rr@{}}
\toprule
Cell type & Post & \makecell{Pre\\+ Post} & \makecell{Pre + Mid\\+ Post} & \makecell{Full\\Post} \\
\midrule
A172 & 8.67 & 19.97 & 33.87 & 35.97 \\
BT474 & 10.59 & 23.36 & 40.84 & 44.36 \\
BV2 & 32.02 & 51.87 & 54.53 & 56.98 \\
Huh7 & 3.71 & 25.24 & 43.36 & 47.13 \\
MCF7 & 10.70 & 21.45 & 39.67 & 42.08 \\
SHSY5Y & 6.01 & 8.31 & 26.33 & 29.10 \\
SKOV3 & 4.31 & 17.41 & 42.78 & 44.97 \\
SkBr3 & 49.32 & 63.96 & 68.47 & 70.59 \\
\bottomrule
\end{tabular*}
\end{table}

\paragraph{Cell-type transfer.}
\textit{Pre} + \textit{Mid} + \textit{Post} improves over direct \textit{Post} for all eight cell types
(\Cref{tab:livecell-types}), with gains of 19.16--39.65 points.
Adding \textit{Mid} to \textit{Pre} + \textit{Post} yields a further 2.66--25.38 points, including
25.38 for SKOV3 and 18.02 for SHSY5Y. With 1\% \textit{Post} GT, the pipeline
comes within 2.10--3.77 points of the full-GT reference across all types.

\Cref{fig:app-transfer-qualitative} illustrates how the staged recipe
improves object completeness and instance separation with limited
human supervision, alongside the full-GT reference.

% END CROSS-TASK APPENDIX 20261121